\documentclass[sigconf]{acmart}
\usepackage{graphicx,subcaption}

\AtBeginDocument{%
  }

\copyrightyear{2023}
\acmYear{2023}
\setcopyright{acmlicensed}\acmConference[SBGames 2023]{22nd Brazilian Symposium on Games and Digital Entertainment}{November 6-9, 2023}{Rio Grande (RS), Brazil}
\acmBooktitle{22nd Brazilian Symposium on Games and Digital Entertainment (SBGames 2023), June 11-September 11, 2023, Rio Grande (RS), Brazil}
\acmPrice{15.00}
\acmDOI{10.1145/3631085.3631231}
\acmISBN{979-8-4007-1627-0/23/06}

\begin{document}

\title[Can we truly transfer an actor's genuine happiness to avatars?]{Can we truly transfer an actor's genuine happiness to avatars? An investigation into virtual, real, posed and spontaneous faces}


\author{Vitor Miguel Xavier Peres, Greice Pinho Dal Molin and Soraia Raupp Musse \\
Email:
\{vitor.peres,greice.molin\}@edu.pucrs.br, soraia.musse@pucrs.br}

\affiliation{%
  \institution{School of Technology}
  \institution{Pontifical Catholic University of Rio Grande do Sul}
  \city{Porto Alegre}
  \country{Brazil}
}

\begin{abstract}
    
“A look is worth a thousand words” is a popular phrase. And why is a simple look enough to portray our feelings about something or someone? Behind this question are the theoretical foundations of the field of psychology regarding social cognition and the studies of psychologist Paul Ekman. Facial expressions, as a form of non-verbal communication, are the primary way to transmit emotions between human beings. The set of movements and expressions of facial muscles that convey some emotional state of the individual to their observers are targets of studies in many areas. 
Our research aims to evaluate Ekman's action units (AU) in datasets of real human faces, posed and spontaneous, and virtual human faces resulting from transferring real faces into Computer Graphics (CG) faces. In addition, we also conducted a case study with specific movie characters, such as She-Hulk and Genius. We intend to find differences and similarities in facial expressions between real and CG datasets, posed and spontaneous faces, and also to consider the actors' genders in the videos. This investigation can help several areas of knowledge, whether using real or virtual human beings, in education, health, entertainment, games, security, and even legal matters. Our results indicate that AU intensities are greater for posed than spontaneous datasets, regardless of gender. Furthermore, there is a smoothing of intensity up to 80\% for AU6 and 45\% for AU12 when a real face is transformed into CG.

\end{abstract}




\maketitle

\section{Introduction}

Expressing emotions is a crucial component of face-to-face communication, as highlighted in a seminal study by psychologist Ekman et al.~\cite{ekman1969repertoire}. As a result, our perception as humans is critical regarding the facial expressions of virtual humans and even robots. Studies on human facial expressions are essential in creating more realistic features for game and movie characters, enhancing user experiences. To this end, performance-driven animations~\cite{Lance1990, Barros2019} have been employed in the game and movie industries to transfer facial motion from real actors to virtual characters. While evaluating such characters typically occurs through movie and game reviews via audience opinions, this paper aims to investigate how the transformation between the real and virtual domains takes place. For instance, is a real human face movement truly represented when transferred to a virtual face? So, the main investigation we intend to pursue in this work concerns the difference/similarities between real and correspondent CG faces.

Numerous researchers have explored facial expression analysis. Barrett et al.~\cite{barrett2019emotional} conducted a literature review of the six categories of emotions, showing how people move their faces during various emotional states and assessing which emotions can be inferred from facial movements. This study is called emotional perception. Park et al.'s~\cite{park:2020} work with real faces has shown that spontaneous smiles have higher intensity in the Upper face region (eyes) than the Lower face (nose and mouth), whereas posed smiles have more intense Lower-face activation. Through the extraction of the Facial Action Coding System (FACS)~\cite{ekman1978}, a comprehensive anatomical taxonomy designed to categorize all facial movements (consisting of 46 distinct Action Unit) and the union of these movements (Action Unit) given a target emotion, it becomes possible to distinguish between posed and spontaneous images by quantifying the relative intensity of Upper and Lower facial expressions~\cite{park:2020}.


According to M{\"a}k{\"a}r{\"a}inen et al.~\cite{makarainen2014exaggerating}, simplifying facial expressions in the transfer from real to Computer Graphics (CG) domains can provide artists with greater freedom to enhance the final facial expression. The authors also believe that reducing realism can decrease emotional intensity in facial expressions, and exaggeration can compensate for this reduction.

In this work, we aim to compare real and correspondent CG faces when expressing happiness, according to ANOVA and Pearson Correlation metrics. Firstly, we want to analyze the discrete activation of AUs in the Upper and Lower face regions, as proposed by Park et al.'s~\cite{park:2020}; secondly, we want to study if the transferring from Real to CG faces changes if the facial expression is Spontaneous or Posed; and finally, we want to investigate if the actors' gender change the way the facial emotions are transferred from real to CG faces. This article seeks to answer the following research questions:

\begin{figure*}[t!] 
    \centering
    \includegraphics[width=1\textwidth]{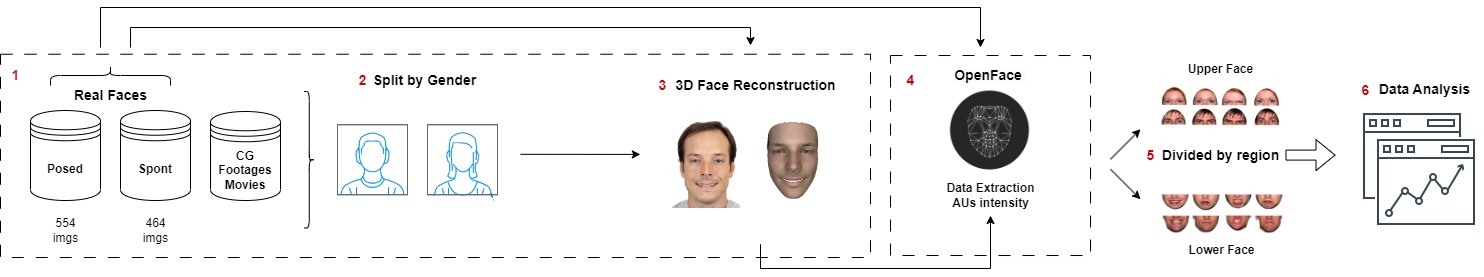}
    \caption{Overview of our method: 
    1) obtaining real faces (Posed and Spontaneous dataset) and CG Characters (detailed in Section~\ref{sec:pipe1} and ~\ref{sec:pipe2}), 
    2) Split all faces by gender,
     3) 3D face reconstruction (Deep3D~\cite{deng2019accurate}, Deca~\cite{DECA:Siggraph2021} and Emoca~\cite{EMOCA:CVPR:2021}) (detailed in Section~\ref{sec:pipe3}), 
     4) Data extraction using OpenFace~\cite{amos2016openface}(detailed in Section~\ref{sec:pipe4}), 
     5) Organize the data between the Upper and Lower regions, and
     6) Data analysis (detailed in Section~\ref{sec:dataAnalysis}).}
    \label{fig:pipeline_pesquisa}
\end{figure*}

\begin{itemize}
   \item \textbf{RQ1:} Are the real faces and their respective correspondences in CG similar considering the activated AU (Lower or Upper parts of the face), the actors' gender (Women or Men), and the type of dataset (Spontaneous or Posed), considering only the happiness emotion?
   \item \textbf{RQ2:} Is there a smooth effect when transforming from Real to CG faces?
   \item \textbf{RQ3:} How do the attributes defined in RQ1 (detailed in Section~\ref{sec:Comparison}) work for existing animated characters from the movie industry?
\end{itemize}

To facilitate a comparison with Park's method~\cite{park:2020}, we have limited our study to the emotion of happiness. This research explores how the Facial Action Coding System (FACS) behaves when analyzing faces by region (Upper and Lower) and across male and female genders. We also tested computer-generated (CG) faces to compare real and virtual images comprehensively. Additionally, we will also compare our results with the work of M{\"a}k{\"a}r{\"a}inen et al.~\cite{makarainen2014exaggerating}, concerning the smoothing expected when transferring real to CG faces. Our work's main contribution is to provide a methodology for analyzing an actor's facial movements and the corresponding animated face to explore the similarities and differences. A possible application could be to guide the needed exaggeration of final expressions when transferring from real to CG faces. This analysis would enable post-processing by artists to more accurately express the facial expressions of virtual characters concerning the real actors' actions. The findings of this research will help guide the creation of virtual characters and their expressions to reflect human emotions with more accuracy.

\section{Related Work}
\label{RelatedWork}

Studies on evaluating facial expressions and Action Units in real and CG faces are the focus of this research. A literature review is done by Barret et al. ~\cite{barrett2019emotional} about the six categories of emotions. Their studies show how people move their faces during different moments of emotions, performing evaluations on which emotions are inferred through facial movements. That study was called Emotional Perception. Park et al.~\cite{park2020differences} explored differences in facial expressions of happy emotion. They performed an analysis to measure changes in the dynamics of the 68 landmark regions. They obtained results indicating that the spontaneous smile is more intense on the upper face than the Lower one, contrary to what happens in faces with a Posed smile. These findings suggest an application of emotion recognition based on landmarks.

The study by Fan et al.~\cite{fan2021demographic} aims to investigate the demographic effects on the expression of facial emotions through the facial action units of happiness indicated by the intensity levels of two distinct facial action units, the Cheek Raiser (AU6) and the Lip Corner Puller (AU12). They statistically analyze the intensity of happiness through the evaluation of each AU and the interaction of three demographic factors, such as gender, age, and race. Their studies indicated a higher intensity of AU12 in women than men. Their research also shows that AI-powered social Rendering can be used to measure the facial expression of happiness by assessing the happiness intensity of AU, to validate the
theory of human emotions into three closely related fields: psychology, anthropology, and social studies.

For Sohre et al.~\cite{sohre2017data}, the study aims to generate a non-parametric face classifier to predict the level of happiness to be perceived. They comment that humans use and expect faces to produce a variety of cues to communicate intent through intonation and emotion. Happiness, for example, is among one of the most basic and important emotions the human face conveys. Since smiles vary widely within and between individuals, creating attractive virtual characters must also exhibit these types of diversity.
In this research, it is observed that facial performance capture, although effective, can involve parameter calibration and requires state-of-the-art equipment, as well as actors who can produce all the desired facial expressions.

Regarding the transferring from real to CG faces, an interesting study in the context of the present work is proposed by M{\"a}k{\"a}r{\"a}inen et al. ~\cite{makarainen2014exaggerating}. The authors believe that simplifying facial expressions, when transferring from the real to the CG domain, gives the artist more freedom to enhance expressions. Just as they believe that facial expressions lose emotional intensity when the level of realism is reduced, exaggeration can compensate for this reduction.

In the present work, we want to answer the questions related to the main research question: \textbf{Are we truly transferring the actor’s genuine happiness to avatars?} For this, we define some metrics that should be used to answer this question and compare our results with the work of Park et al.~\cite{park2020differences}, regarding the Upper and Lower parts of the face; with the study by Fan et al.~\cite{fan2021demographic} which indicated a higher intensity of AU12 in women than men, during happiness emotion; and finally, with the work of by M{\"a}k{\"a}r{\"a}inen et al. ~\cite{makarainen2014exaggerating} to evaluate the smoothness of emotion when transferring from real to CG faces.

\begin{figure*}[t!]
    \centering
    \includegraphics[width=1\textwidth]{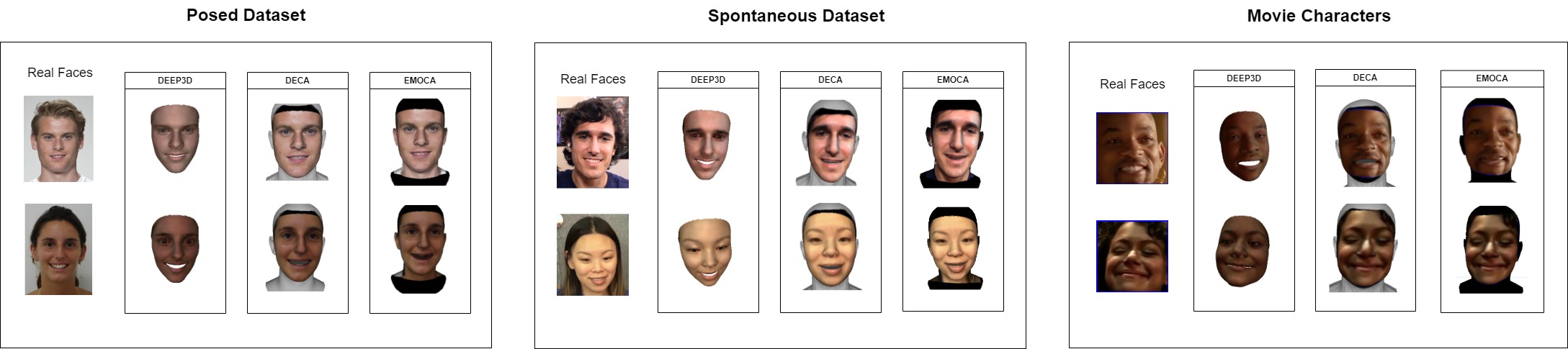}
    \caption{Visual representation of datasets used in this paper, given the Posed and Spontaneous datasets. And the CG characters' faces footage from Aladdin (2019) and She-Hulk (2022).}
    \label{fig:dataset_show}
\end{figure*}

\section{Data Acquisition and Generation}
\label{OurMethodology}

In literature, there are two types of datasets of real facial expressions: Posed and Spontaneous~\cite{motley1988facial}. In addition to the analysis performed in the real faces datasets, as it is going to be discussed in the next sections, we also transferred real to CG faces using the Deep3D~\cite{deng2019accurate}, Emoca~\cite{EMOCA:CVPR:2021} and Deca~\cite{DECA:Siggraph2021} software, generating two more datasets (CG Posed and CG Spontaneous). We extract the Action Units for all studied datasets using OpenFace~\cite{amos2016openface}, a toolkit capable of facial landmark detection, head pose estimation, facial action units recognition, and eye-gaze estimation.

The overview of our method is illustrated in Figure~\ref{fig:pipeline_pesquisa}. Firstly, we obtained the real faces from existing datasets; secondly, we split such datasets using gender notations. The next phase is responsible for transferring the real faces into CG faces through the Deep3D\cite{deng2019accurate}, Emoca~\cite{EMOCA:CVPR:2021} and Deca~\cite{DECA:Siggraph2021}. Then, all datasets (Real and CG) extract AUs using OpenFace~\cite{amos2016openface}. Finally, the datasets are again separated regarding the activation of AUs (Lower and Upper parts of the face). The following sections detail the used data.


\subsection{Real Faces Datasets}
\label{sec:pipe1}

In this paper, we use four Posed datasets (CFD~\cite{Ma:2015} and CFD-INDIA~\cite{Lakshmi2021TheIF}, FEI~\cite{THOMAZ2010902} and London~\cite{debruine2017face}) and two Spontaneous (Reacts~\cite{peres2021towards} and DISFA~\cite{disfa:2013}) datasets, where we choose only the footages that display the emotion Happiness. The Chicago Face Database (CFD)~\cite{Ma:2015} and CFD-INDIA~\cite{Lakshmi2021TheIF}, consist of 587 and 142, respectively, high-resolution standardized photographs of Black and White males and females between the ages of 18 and 40 years. The CFD~\cite{Ma:2015} was expanded to include multiracial faces and Indian faces photographed CFD-INDIA~\cite{Lakshmi2021TheIF} and presents the equal distribution of samples for men and women.

FEI~\cite{THOMAZ2010902} is a dataset of frontal and pre-aligned images of a face database maintained by the Department of Electrical Engineering of FEI. It contains 200 faces (100 men and 100 women), where each subject has two frontal images (one with a neutral expression and the other with a smiling facial expression) and is between 19 and 40 years old with distinct appearance, hairstyle, and adornments. London~\cite{debruine2017face} makes available 102 adult faces 1350x1350 pixels in full color. Self-reported age, gender, and ethnicity are included, as are attractiveness ratings (on a scale of 1-7, from "much less attractive than average" to "much more attractive than average").

Reacts~\cite{peres2021towards} presents a method of creating a spontaneous facial expression dataset from YouTube reaction videos. From 18,400 images (frames) extracted from the YouTube reaction videos, the authors selected 732 images, with the six basic emotions and 80\% as the threshold minimum in the VGG16 accuracy. DISFA~\cite{disfa:2013} contains approximately 130,000 annotated frames from 27 subjects. For every frame in this dataset, the intensity of the 12 action units was manually annotated on a six-point ordinal scale (0 and 5 increasing intensity levels). Also, we only use faces that represent Happiness emotion.  

To ensure that all analyzed faces in the datasets include the corresponding action units (AU) for happiness emotions, we employ the OpenFace~\cite{amos2016openface} Confidence variable as a threshold. This variable provides an estimate of the reliability of the current landmark detection. We only consider faces with a confidence level of 90\% or higher in the emotion classification, ensuring their reliability for analysis. Finally, we utilized a dataset consisting of 554 images of Posed faces, with 281 images of men and 273 images of women. Additionally, we included 464 images of Spontaneous faces, comprising 247 images of men and 217 images of women. 

\subsection{Character's Faces Footages}
\label{sec:pipe2}

To complement our analysis, we collected images from two popular Hollywood movie characters: Genius from Aladdin (2019) and She-Hulk from the Marvel TV series (2022). The collection of these characters' faces in CG was based on detecting their faces in the YouTube video trailers\footnote{Genius - https://www.youtube.com/watch?v=zWgBZ7erqDo}\footnote{She-Hulk - https://www.youtube.com/watch?v=a-oeACwhVI0}. Figure~\ref{fig:dataset_show} illustrates the footage collected in this paper. As the main goal is to analyze only faces that express the emotion of happiness, we use the pyFeat~\cite{pyfeat:2021} tool, which classifies all the faces collected from the videos among the six basic emotions since we do not have the emotions' annotation of the faces extracted from each film.

The collections contained images of the human form (real faces) of the characters played by actress Tatiana Maslany and Will Smith, respectively, and the CGI faces of the characters reproduced by the animators. This allowed us to compare faces animated by the studio and faces produced by 3D reconstruction models.



\subsection{Transforming Real Faces to CG Faces}
\label{sec:pipe3}

In addition to the real faces, we are interested in studying correspondent CG faces. Therefore, we propose reconstructing CG faces based on real ones using the three following facial reconstruction models: Deep3D~\cite{deng2019accurate}, Deca~\cite{DECA:Siggraph2021} and Emoca\cite{EMOCA:CVPR:2021}. 

Deep3D~\cite{deng2019accurate} proposed a novel multi-image face reconstruction aggregation method using Convolutional Neural Rendering without any explicit label. Figure~\ref{fig:dataset_show} illustrates the process proposed in this paper. Each dataset containing real faces is transformed to CG, even for the media characters She-Hulk and Genius. Deca~\cite{DECA:Siggraph2021} (Detailed Expression Capture and Animation) is a model that learned from in-the-wild images with no paired 3D supervision. Deca reconstructs a 3D head model with detailed facial geometry from a single input image. Emoca~\cite{EMOCA:CVPR:2021} (Emotion Capture and Animation) is a neural network that learns an animatable face model from in-the-wild images without 3D and introduces a novel perceptual emotion consistency loss that encourages the similarity of emotional content between the input and rendered reconstruction supervision. 



\subsection{Data Processing using OpenFace}
\label{sec:pipe4}

Once the datasets are organized as illustrated in Figure~\ref{fig:pipeline_pesquisa}, we use OpenFace~\cite{amos2016openface} to find the AU intensities in each face (real or CG). OpenFace is an open-source facial recognition software that uses deep learning to determine facial landmarks, head poses, and Action Units (AU). As presented in literature~\cite{ekman1978,Keltner:2019}, the six basic emotions are described as a set of activated AU (Action Units). However, as mentioned before, in our paper, we are analyzing only \textit{Happiness}, which is compounded by AU06 (cheek raiser) and AU12 (lip corner puller)~\cite{Ghayoumi}. For all the analyzed faces, Real or CG, Posed or Spontaneous, we split them into new groups of features. We also consider the face gender and the upper (AU6) and lower face parts (AU12) activated in the expressions. 

The Convolutional Expert Constrained Local Model (CE-CLM)~\cite{8265507} and Multi-task Convolutional Neural Network (MTCNN)~\cite{7553523} model configurations were set for detecting facial landmarks and detecting faces in OpenFace. The following section details the data analysis proposed in this work.

\section{Data Analysis}
\label{sec:dataAnalysis}

This section includes a quantitative comparison of 3D reconstruction techniques (ANOVA) concerning the type of dataset (Posed and Spontaneous), image domain (Real and CG), and actor's gender (Men and Women). We also use the Cosine Similarity and Structural Similarity Index (SSIM) metrics to evaluate the behavior of faces generated by 3D reconstruction models when compared with their respective real faces. In addition, we present a case study of two media characters from Blockbuster's productions: Genius (Aladdin - 2019) and She-Hulk (She-Hulk: Attorney at Law).

\begin{figure*}[!t]
    \minipage{0.32\textwidth}
    \includegraphics[width=\linewidth]{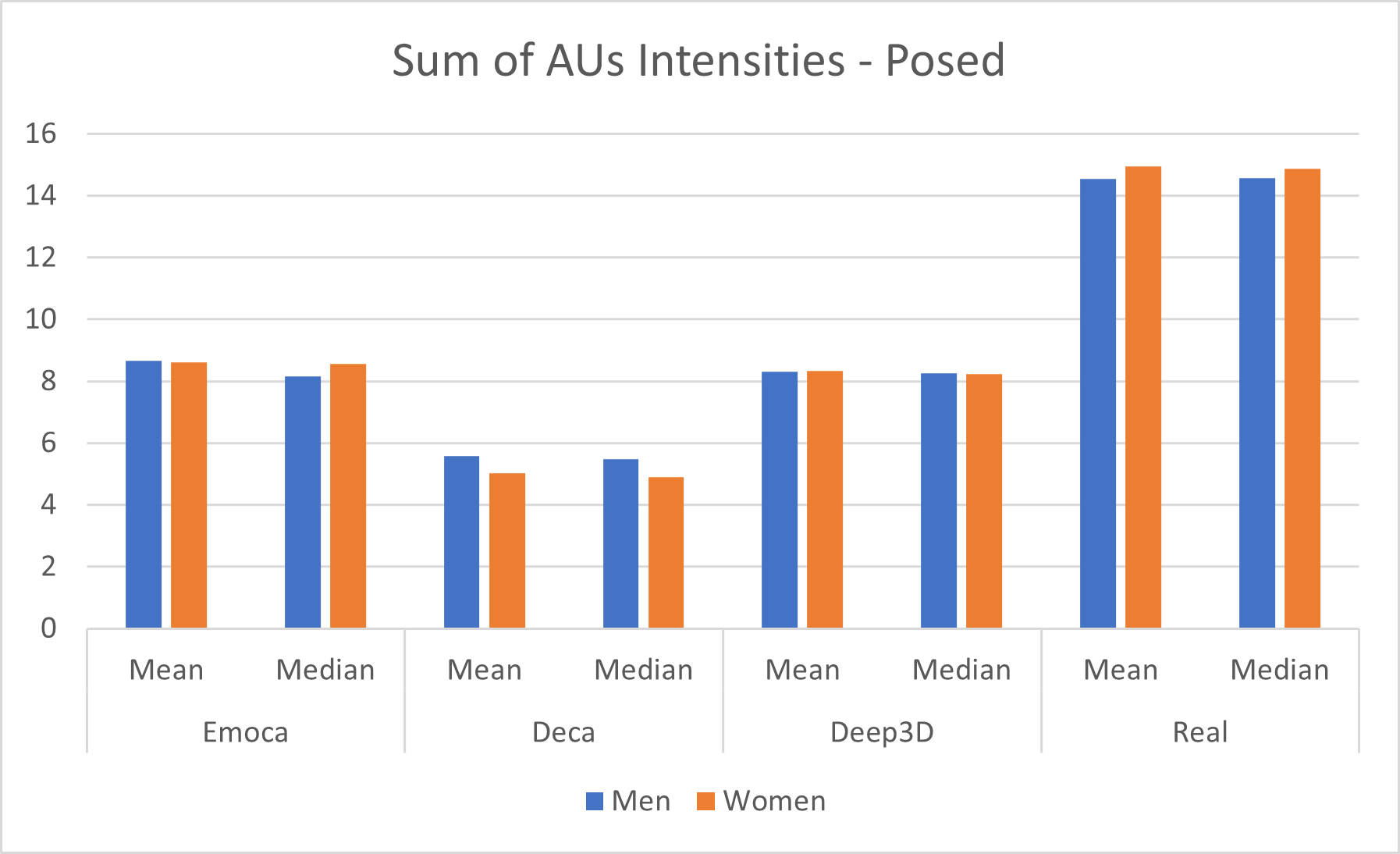}
    \caption{Mean and Median sum of all Intensities of Actions Units, generated in each model and the real one. Separated by genre and for the posed dataset.}\label{fig:awesome_image1}
    \endminipage\hfill
    \minipage{0.32\textwidth}
    \includegraphics[width=\linewidth]{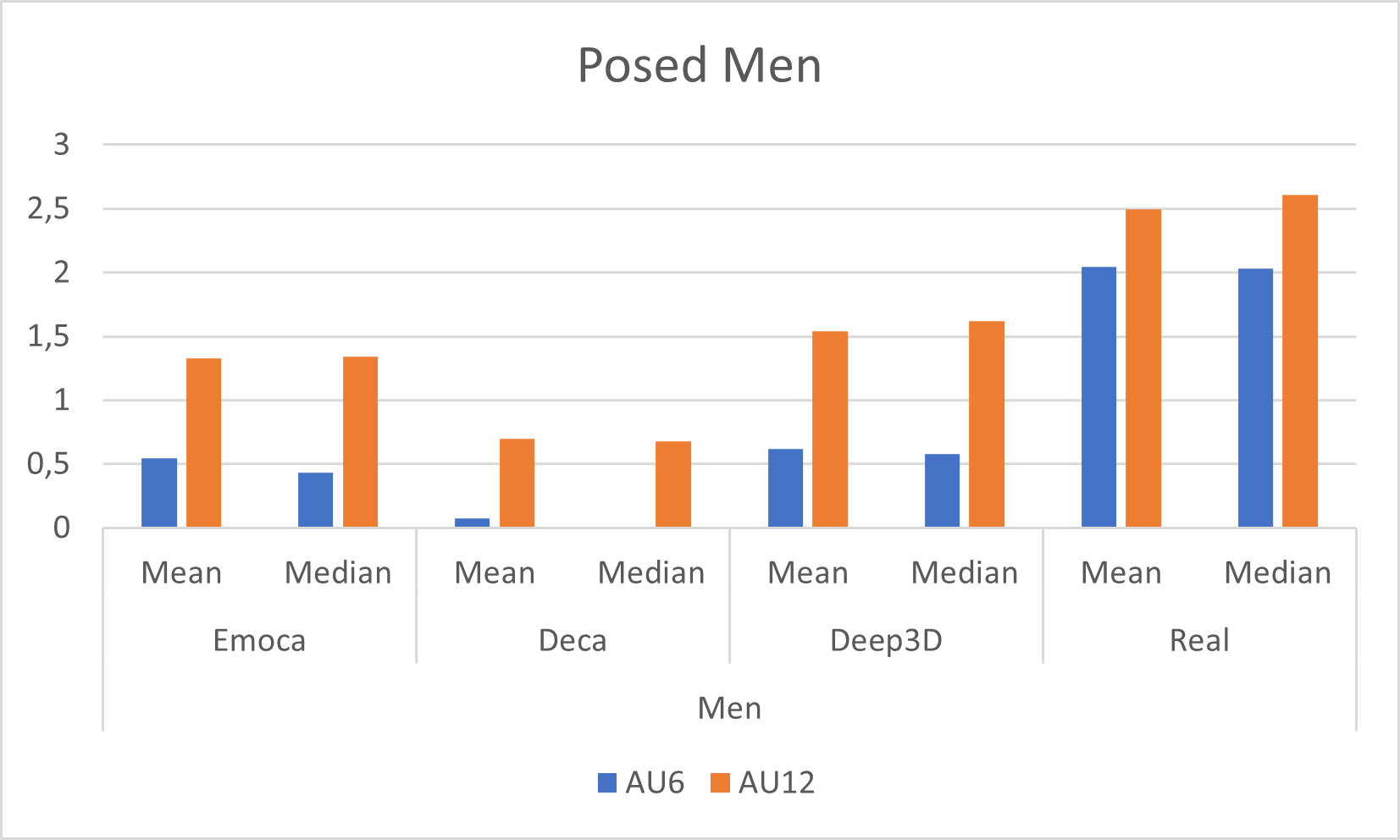}
    \caption{Mean and median of the intensity of the Actions units that compound the expression of happiness (AU6 and AU12), generated in each model and the real one, from dataset posed men.}\label{fig:awesome_image2}
    \endminipage\hfill
    \minipage{0.32\textwidth}%
    \includegraphics[width=\linewidth]{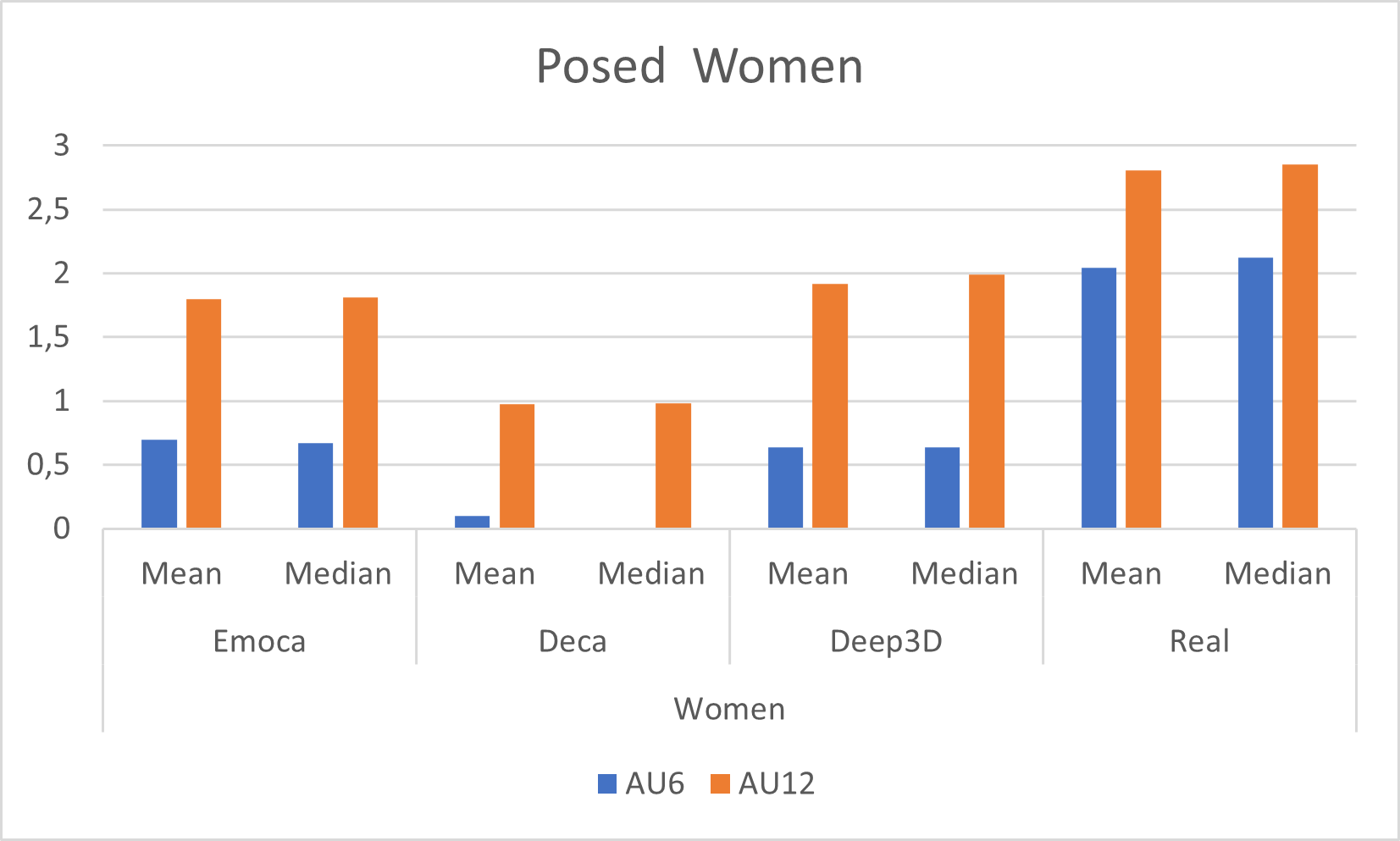}
    \caption{Mean and median of the intensity of the Actions units that compound the expression of happiness (AU6 and AU12), generated in each model and the real one, from dataset posed women.}\label{fig:awesome_image3}
    \endminipage
\end{figure*}

\subsection{Quantitative Comparison of 3D Models}

To evaluate the reconstructed 3D faces, we employ two metrics on 2D images to compute the similarity between the input images (real faces) and the rendered images (3D faces). We use the Cosine Similarity (Equation~\ref{eq:eqt1}) of features vectors (17 Facial action units)~\cite{cai2021multi}, which are extracted from OpenFace~\cite{amos2016openface}, where $\left\| A\right\|$ represents the length of the vector A, $\theta$ denotes the angle between $A$ and $B$, and ‘$.$’ denotes the dot product operator.

\begin{equation}
    \label{eq:eqt1}
    \cos (\theta ) =   \dfrac {A \cdot B} {\left\| A\right\| x \left\| B\right\|.} 
\end{equation}

The Structural Similarity Index (SSIM)~\cite{lin2020towards} was used, where $\mu_x$, $\mu_y$, $\sigma_x$, $\sigma_y$ and $\sigma_xy$ are the local means, standard deviations and cross-covariance for images $x, y$. SSIM enables us to evaluate results at the pixel level. Equation~\ref{eq:eqt2} defines the SSIM:

\begin{equation}
    \label{eq:eqt2}
    SSIM(x,y) = \dfrac{ (2\mu_x\mu_y + c_1)(2\sigma_xy + c_2) }{ (\mu_x^2 + \mu_y^2 + c_1)(\sigma_x^2 + \sigma_y^2 + c_2) }
\end{equation}

The SSIM's goal is to assess perceptual image quality traditionally attempted to quantify the visibility of errors between a distorted image and a reference image using a variety of known properties of the human visual system~\cite{nilsson2020understanding, 1284395}. The Human visual perception can identify structural information from a scene and hence identify the differences between the information extracted from a reference and a sample scene~\cite{1284395}. The quantitative comparison in shown in Tables~\ref{tab:comp1} and ~\ref{tab:comp2}.

\begin{table}[h]
    \centering
    \begin{tabular}{ c c c c }
     \hline Dataset & Deep3D~\cite{deng2019accurate} & Emoca~\cite{EMOCA:CVPR:2021} & Deca~\cite{DECA:Siggraph2021} \\ \hline \hline
     Posed Men & \textbf{0.6432} $\uparrow$ & 0.602 & 0.4587 \\
     Posed Women & \textbf{0.7430} $\uparrow$ & 0.7412 & 0.5839 \\
     Spont Men & \textbf{0.6574} $\uparrow$ & 0.6340 & 0.5305  \\
     Spont Women & 0.6471 & \textbf{0.7108} $\uparrow$ & 0.5542  \\ \hline
     She-Hulk & 0.5843 & \textbf{0.6288} $\uparrow$ & 0.4779  \\
     Genius & \textbf{0.6516} $\uparrow$ & 0.4913 & 0.4022  \\ \hline
    \end{tabular}
    \caption{Quantitative comparison of mean cosine similarity features (Facial Action Units) for each model 3D reconstructed through the Real faces. "Spont" states for Spontaneous. The arrow up $\uparrow$, represents the highest achieved values of Cosine Similarity.}
    \label{tab:comp1}
\end{table}

\vspace{-2em}

\begin{table}[h]
    \centering
    \begin{tabular}{ c c c c }
     \hline Dataset & Deep3D~\cite{deng2019accurate} & Emoca~\cite{EMOCA:CVPR:2021} & Deca~\cite{DECA:Siggraph2021} \\ \hline \hline
     Posed Men & \textbf{58.8\%} $\uparrow$ & 46.4\% & 21.0\% \\
     Posed Women & 43.9\% & 18.9\% & \textbf{56.3\%} $\uparrow$\\
     Spont Men & 30.0\% & 24.0\% & \textbf{34.3\%} $\uparrow$\\
     Spont Women & \textbf{42.3\%} $\uparrow$ & 24.3\% & 31.1\%\\ \hline
     She-Hulk & \textbf{24.9\%} $\uparrow$ & 20.5\% & 19.5\% \\
     Genius & \textbf{39.8\%} $\uparrow$ & 17.4\% & 18.9\% \\ \hline
    \end{tabular}
    \caption{Quantative comparison of mean Structural Similarity Index (SSIM), for each model 3D reconstructed through the Real faces. "Spont" states for Spontaneous. The arrow up $\uparrow$, represents the the highest achieved levels of SSIM.}
    \label{tab:comp2}
\end{table}

The results indicate that the mean of Cosine Similarity between real and CG faces was higher when reconstructing Posed (both genders), Spontaneous (Men), and Genius, using Deep3D. Emoca obtained higher values for Spontaneous (Women) and She-Hulk. Notably, the Deca model obtained the lowest results in all datasets. Considering the SSIM value, we observe that Deep3D faces are closer when processing the Posed (Men), Spontaneous (Women), and She-Hulk and Genius in pixel levels.  

The process of 3D reconstruction shows the effect of decreasing intensities AUs intensities when compared to the real faces, as shown in Figures~\ref{fig:awesome_image1},~\ref{fig:awesome_image2} and~\ref{fig:awesome_image3}. Figure~\ref{fig:awesome_image1} displays the mean (and median) sum of all Action Units (AUs) generated by OpenFace for each set of faces generated for each 3D model and the real faces from the posed dataset. In Figures~\ref{fig:awesome_image2} and ~\ref{fig:awesome_image3}, we present the mean difference (and median) in AU intensities for the emotion of happiness (AU6 and AU12) in the posed dataset, encompassing both genres. Based on the SSIM and Cosine Similarity metrics and the AU intensities, which exhibited greater proximity compared to the real faces, we proceeded to the following analysis using the data generated by the Deep3D model.

\subsection{Comparative Analysis}
\label{sec:Comparison}

\begin{figure*}[t!]
    \centering
    \begin{subfigure}[b]{0.4\linewidth}        
        \centering
        \includegraphics[width=\linewidth]{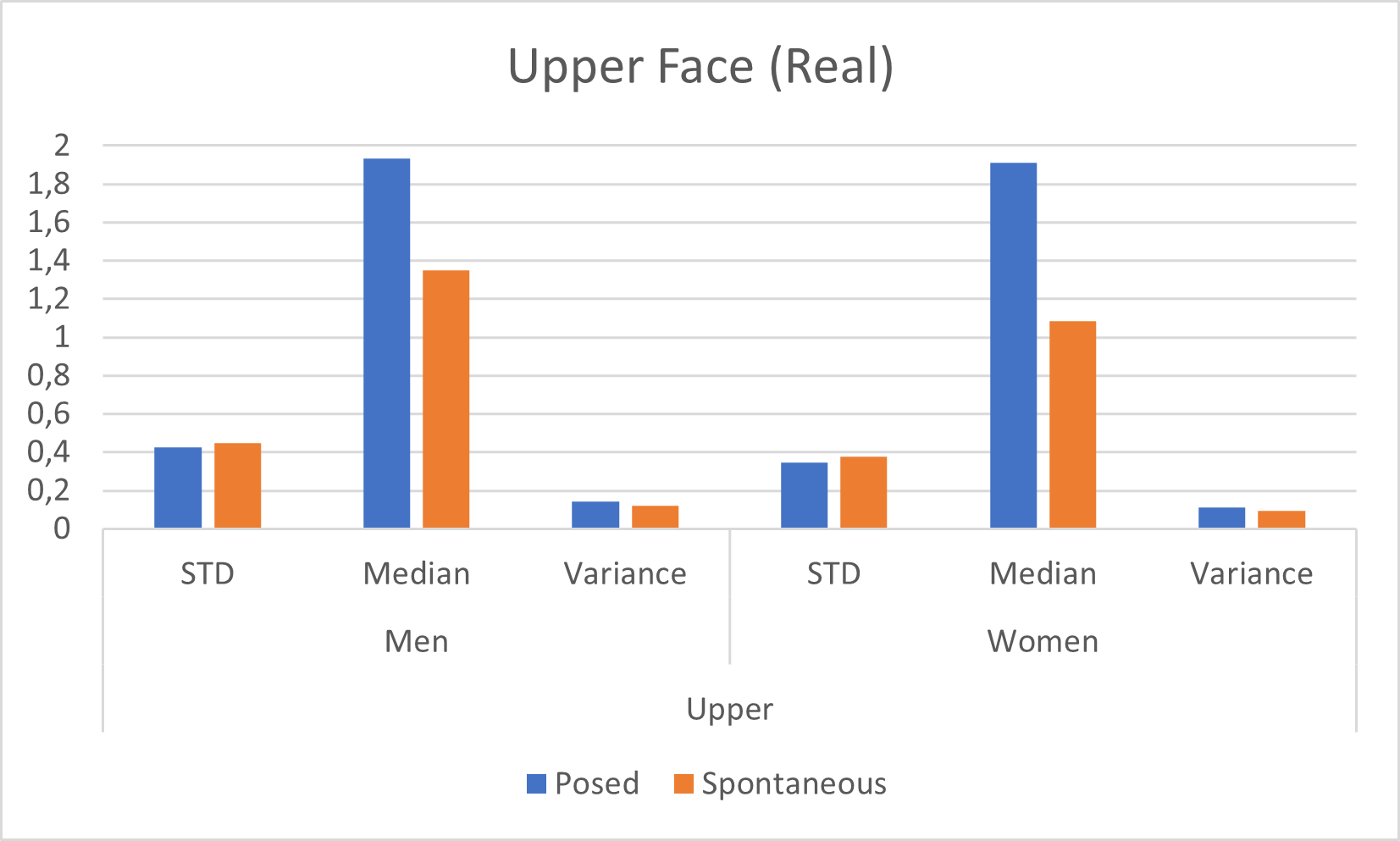}
        \caption{Upper Face (AU6) Real}
        \label{fig:A}
    \end{subfigure}
    \begin{subfigure}[b]{0.4\linewidth}        
        \centering
        \includegraphics[width=\linewidth]{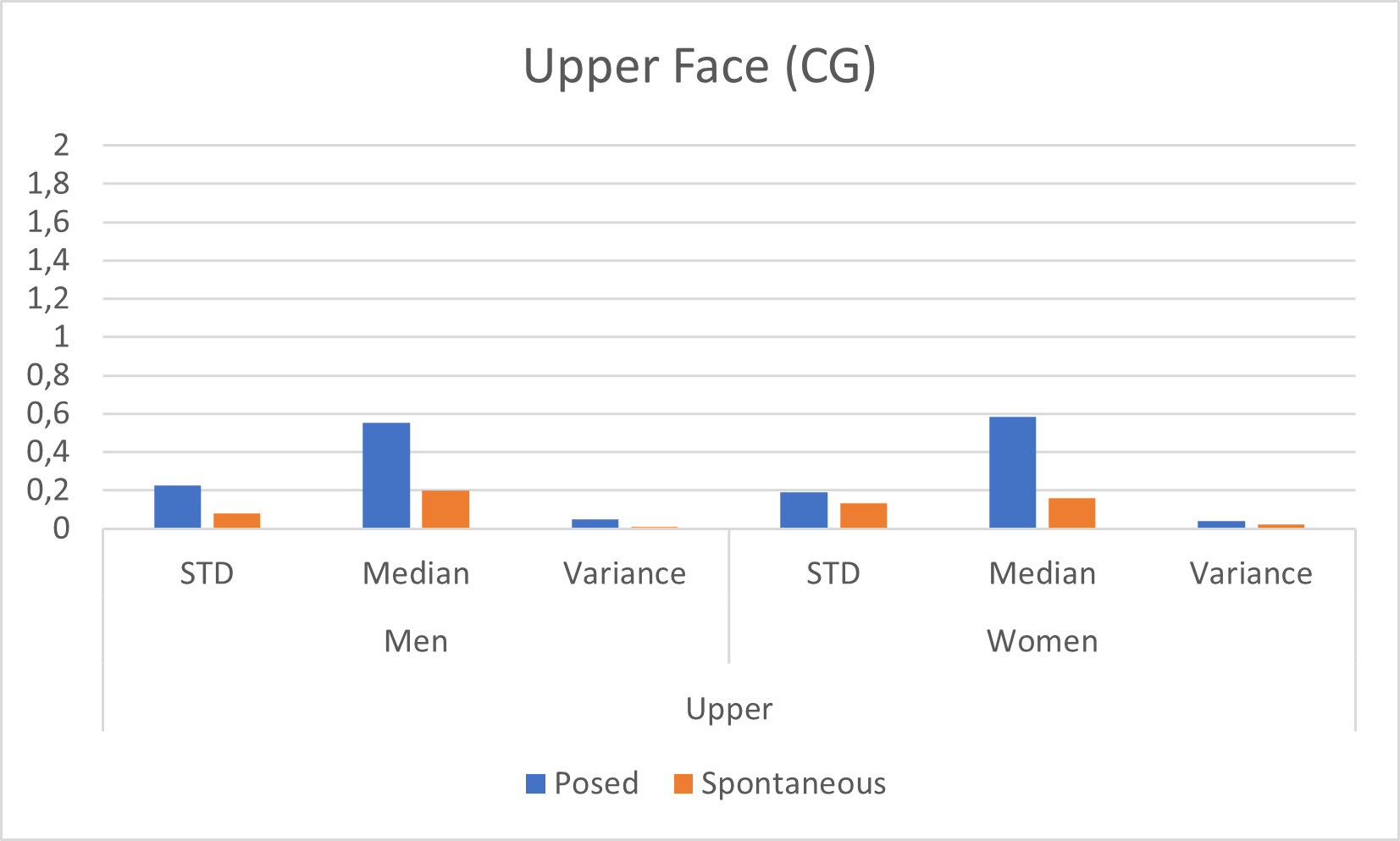}
        \caption{Upper Face (AU6) CG}
        \label{fig:B}
    \end{subfigure}

    \begin{subfigure}[b]{0.4\linewidth}        
        \centering
        \includegraphics[width=\linewidth]{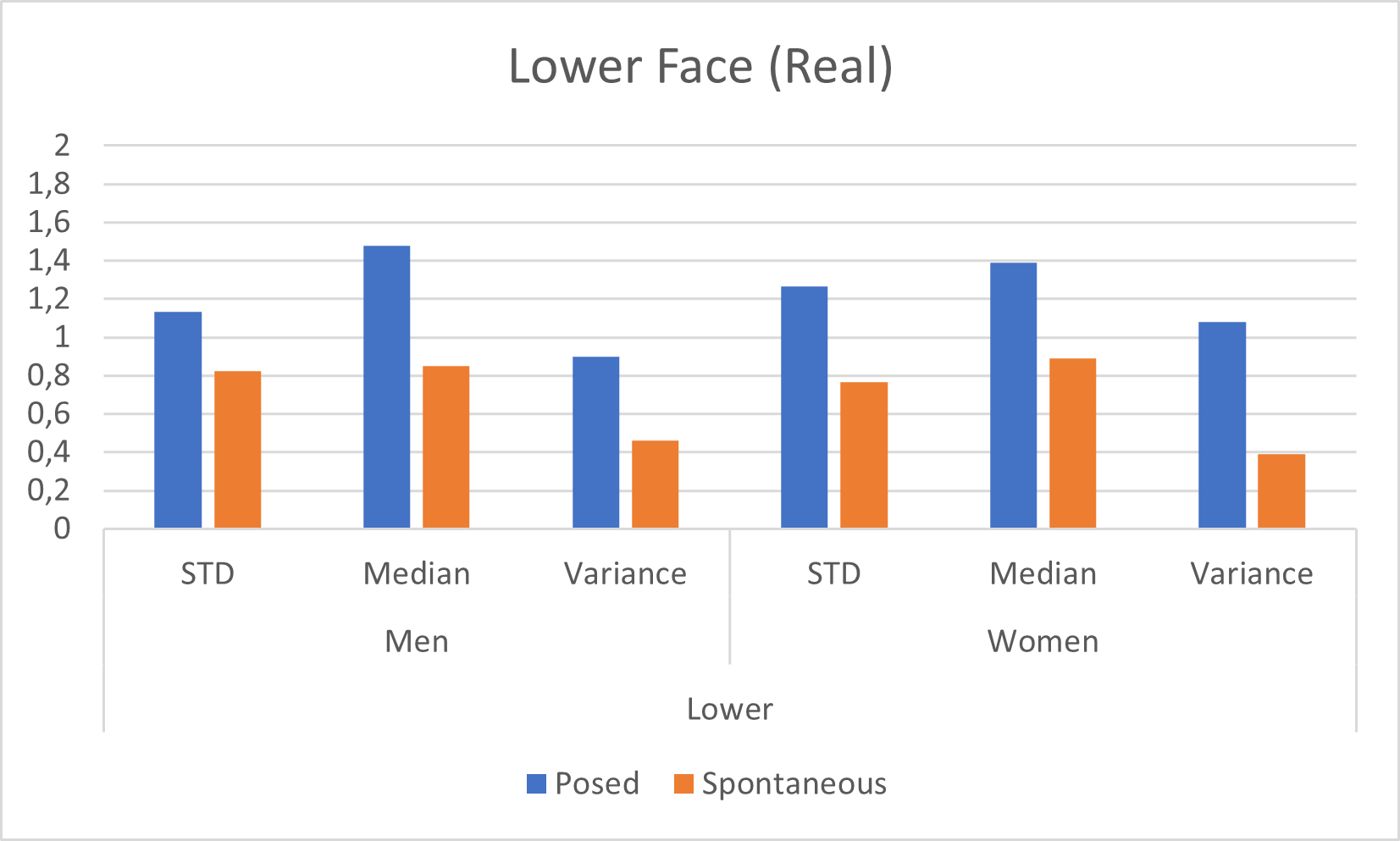}
        \caption{Lower Face (AU12) Real}
        \label{fig:C}
    \end{subfigure}
    \begin{subfigure}[b]{0.4\linewidth}        
        \centering
        \includegraphics[width=\linewidth]{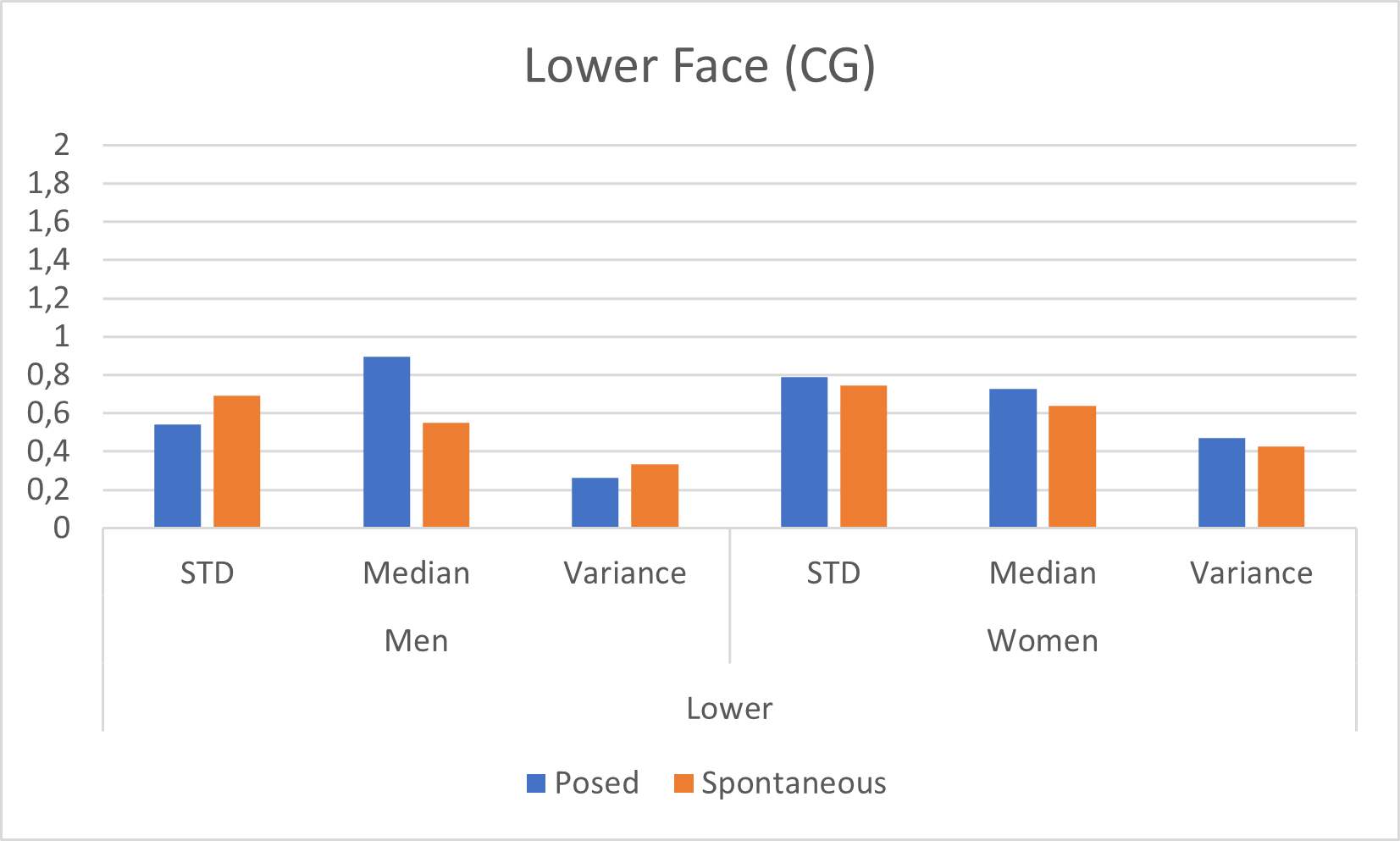}
        \caption{Lower Face (AU12) CG}
        \label{fig:D}
    \end{subfigure}
    
    \caption{Comparisons between Posed and Spontaneous happy faces median, variance, and standard-deviation intensities from Upper (AU6) and Lower (AU12) Faces. This is presented in addition to the actors' gender and Real/CG image features. 
    }
    \label{fig:comparative1}
\end{figure*}

The statistical analysis ANOVA was computed with the intensity values from AU6 (Upper) and AU12 (Lower), as shown in Figure~\ref{fig:comparative1}. In such a figure, we present the median, standard deviation, and variance of activated AU6 and AU12 in the faces included in our datasets. In all statistical analyses, we used 5\% of significance level (\textit{Factorial ANOVA} and \textit{Tukey HSD} tests). The independent variables include Gender (Men x Women), Image domain (Real x CG), and Dataset (Posed and Spontaneous). The dependent variables are AU6 and AU12 intensity, as shown in Table~\ref{tab:anova_1}. 

\begin{table*}[h!]
    \small
    \centering

        \begin{tabular}{l c c c r}
        \hline
            \textbf{Factor} & \textbf{Sum of squares} & \textbf{df} & \textbf{F} & \textbf{$f$-value}  
            \\ [0.5ex]
         \hline
           
            \multicolumn{5}{l}{(a) Three-way ANOVA on AU6 intensity} \\ \hline

              Gender & 0.31371 & 1.0 & 0.92145 & 0.33721 \\ 
              Domain & 1039.04742 & 1.0 & 3051.94215 & \textbf{$<$ 0.00000} 
              \\
              Dataset & 59.78617 & 1.0 & 175.60695 & \textbf{$<$ 1.6753e-38} 
              \\
              Gender x Domain & 2.66359 & 1.0 & 7.82362 & \textbf{$<$ 0.0052} 
              \\
              Gender x Dataset & 0.11457 & 1.0 & 0.33653 & 0.5619 
              \\
              Domain x Dataset & 0.14343 & 1.0 & 0.42128 & 0.5164 
              \\
              Gender x Domain x Dataset & 3.32974 & 1.0 & 9.78029 & \textbf{$<$ 0.0017} 
              \\ \hline

            \multicolumn{5}{l}{(b) Three-way ANOVA on AU12 intensity} \\ \hline
         
              Gender & 7459.18140 & 1.0 & 14965.25303 & \textbf{$<$ 4.7764e-18} 
              \\
              Domain & 38.07696 & 1.0 & 76.39329 & \textbf{$<$ 1.7164e-189} 
              \\
              Dataset & 535.61834 & 1.0 & 1074.60371 & \textbf{$<$ 3.0268e-60} 
              \\
              Gender x Domain & 142.87125 & 1.0 & 286.64063 & 0.6078 
              \\
              Gender x Dataset & 0.13135 & 1.0 & 0.26353 & \textbf{$<$ 0.0360} 
              \\
              Domain x Dataset & 2.19434 & 1.0 & 4.40247 & \textbf{$<$ 0.0005} 
              \\
              Gender x Domain x Dataset & 0.16744 & 1.0 & 0.33592 & 0.56226 
              \\ \hline
              
         \hline
           
        \end{tabular}
   
    \caption{Three-way (Gender x Domain x Dataset) ANOVA of the main effect and the interaction effect on (a) AU6 intensity and (b) AU12 intensity, where gender is men/women, the Domain is CG/real and Dataset is posed/spontaneous. The significant variables and their corresponding \textit{p}-values are highlighted in bold. \textit{p}-values $<$ 0.05 are considered statistically significant.}
    \label{tab:anova_1}
\end{table*}

\sloppy 

Regarding the AU6, the intensity values are presented in Figure~\ref{fig:comparative1} (a) and (b), respectively, in real and CG faces. Specifically, we observed that the median intensities for both Posed and Spontaneous datasets are similar between genders (Men/Women) in CG and Real images, with higher values for Posed datasets than Spontaneous ones. These findings suggest no significant differences in AU6 intensity between men and women. This is supported by the results of the three-way ANOVA (Table~\ref{tab:anova_1}), which indicate that the Gender factor (F=0.9214, $f$-value=0.3372) has a low significant effect. On the other hand, we observed significant results in the domain (CG/Real) and dataset (Posed/Spontaneous) features as Domain factor (F=3051.94215, $f$-value=0.0) and Dataset factor (F=175.60695, $f$-value=1.6753e-38). 
In addition, the interactions containing Gender x Domain (F=7.82362, $f$-value= 0.0052) and Gender x Domain x dataset (F=9.78029, $f$-value=0.0017) were also evaluated. For cases of significant interactions, we performed the post-hoc Tukey test to assess the intensity of the AU6, which shows that, out of 6 gender and domain interactions, the $p$-value was considered below $0.05$, with only two groups having $p$-adj with similar means, not influencing the $p$-value of the Tukey test.

The median intensity of AU12 expressions (Figure~\ref{fig:comparative1} (c) and (d)) is higher in Posed datasets, for both genders, compared to the Spontaneous dataset, as observed also in AU6. In this case, as shown in Table~\ref{tab:anova_1}, the factors Gender (F=14965.25303, $f$-value=4.7764e-18), Domain (F=76.39329, $F$-value=1.7164e -189) and Dataset (F=1074.60371, $f$-value=3.0268e-60) are significant in the AU12 analysis, while the interactions indicate significant values only in Gender x Dataset (F=0.26353, $f$-value= 0.0360) and Domain x Dataset (F=4.40247, $f$-value=0.0005). Tukey's post-hoc test was performed for cases of significant interactions to assess the intensity of the AU12, which shows that, out of 6 gender and domain interactions, the $p$-value was considered below 0.05. There was no interaction with $p$-adj greater than $0.05$, indicating that this interaction influences this $p$-value. For the Domain x Dataset interaction, we have a similar situation.

\begin{figure*}[h]
    \centering
    \subfloat[\centering Pearson Correlation by Region in Posed Datasets.]{{\includegraphics[width=3in]{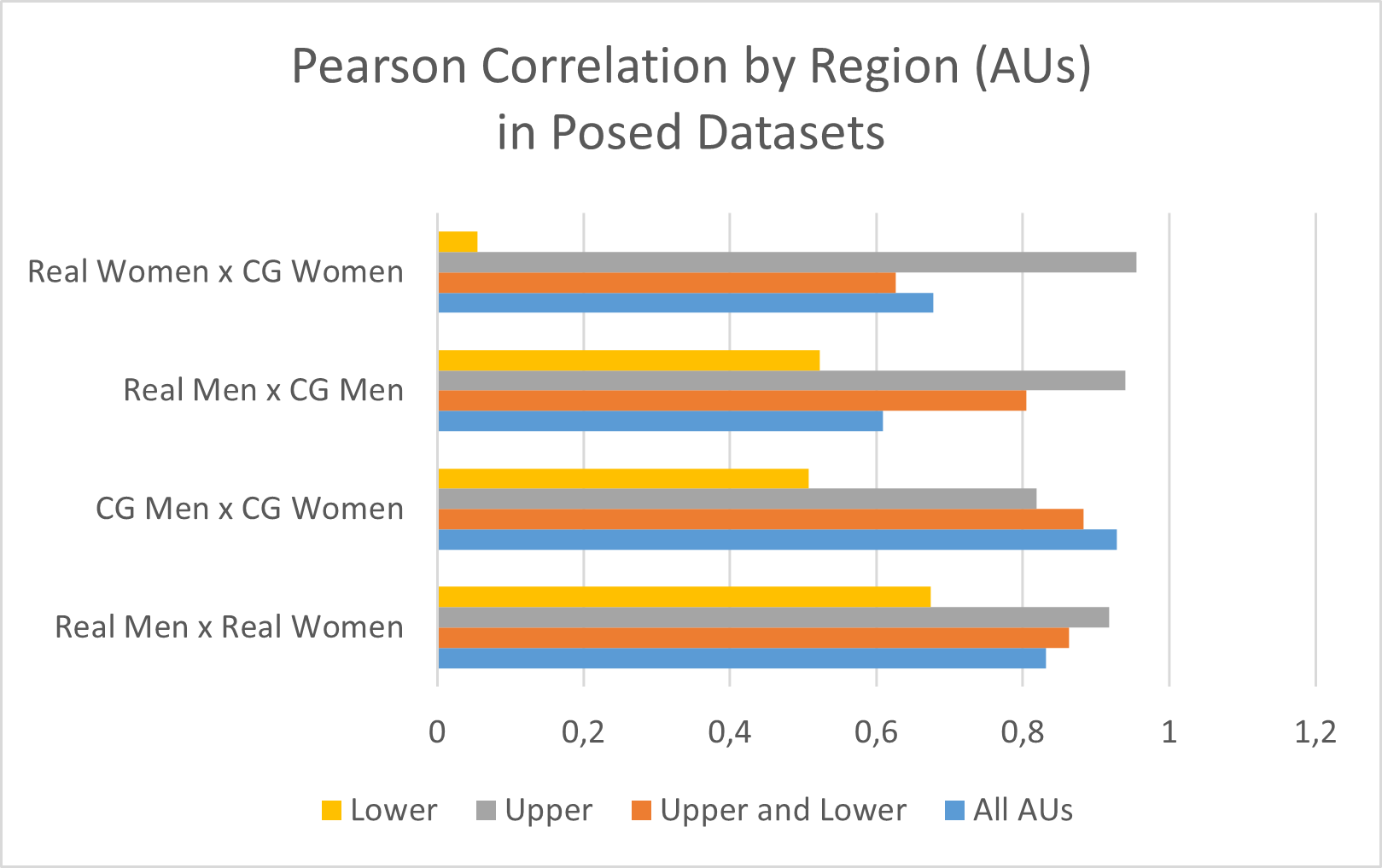} }}%
    \qquad
    \subfloat[\centering Pearson Correlation by Region in Spontaneous Datasets.]{{\includegraphics[width=3in]{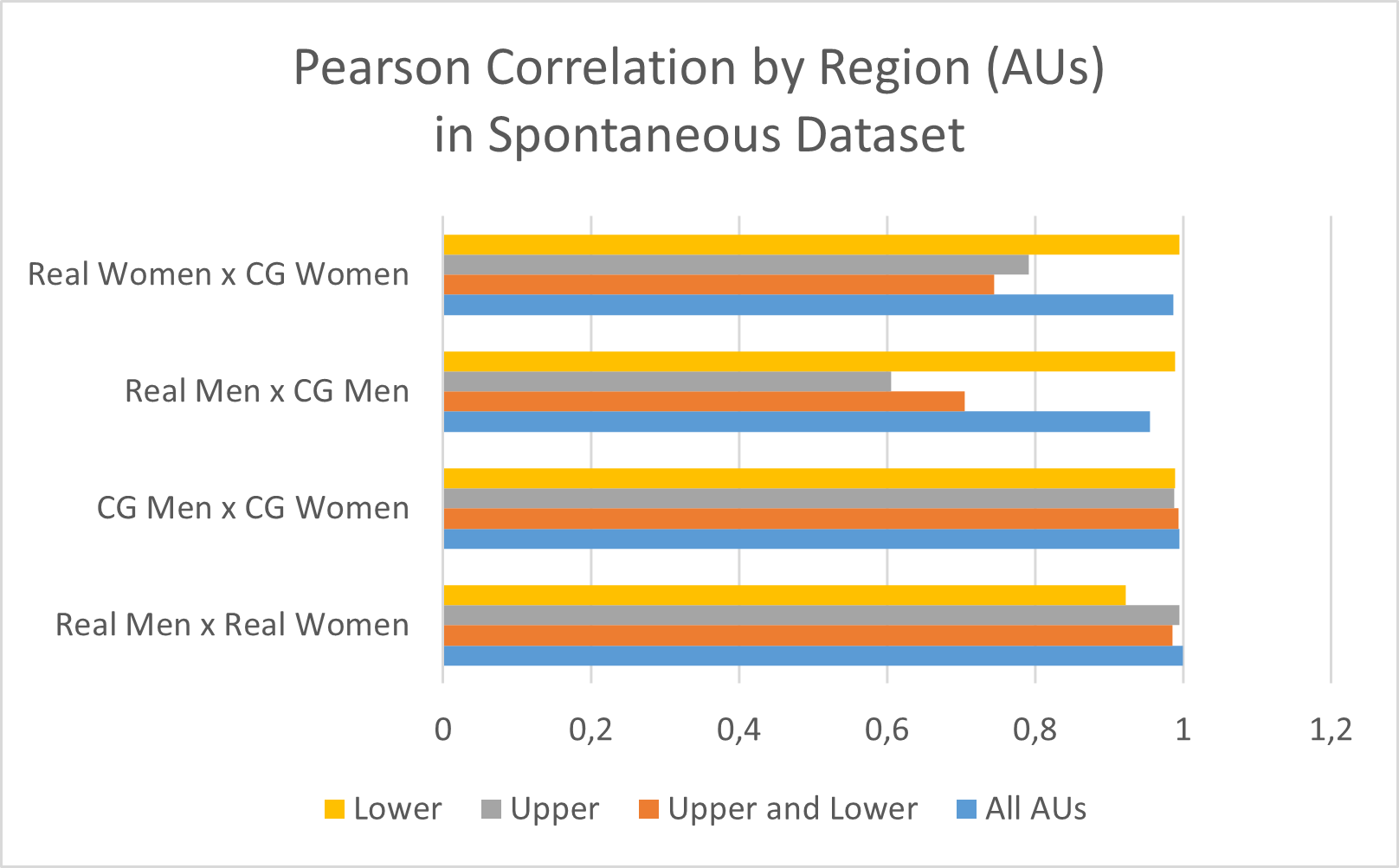} }}%
    \caption{Pearson’s correlations regarding the AU intensity values for AU6, AU12, the both of them and all AU.}
    \label{fig_pearson_correlation_region_AU}
\end{figure*}

In addition to the ANOVA analysis, Figure~\ref{fig_pearson_correlation_region_AU} displays the Pearson correlation analysis between the combination of AU intensities (AU6-Upper, AU12-Lower, AU6/AU12, and All AUs). On the left (Figure~\ref{fig_pearson_correlation_region_AU} (a)), we note that the Posed datasets with Real faces exhibited the strongest correlation in the AU06 (Upper) when we analyzed both genders and domains (Real and CG). Upon analyzing the face's Lower region (AU12) separately, a weak correlation is explicitly observed in the women factor when comparing Real with CG face. This weak correlation is not observed in the other combinations. So, \textbf{we can observe that the Upper part of the face better correlates with Real and CG actors than the Lower part for posed datasets.}

In Figure~\ref{fig_pearson_correlation_region_AU} (b), we noted that the AUs activated in the Spontaneous datasets present a strong correlation between the combination of both Genders and in Real and CG faces (CG men x CG women and Real men x Real women). It is important to note that there is an inversion of correlations observed when comparing AU6 and AU12 between Posed and Spontaneous faces. Specifically, AU12 presents higher correlations in all compared groups in Spontaneous datasets, while AU6 correlates better with the Posed datasets. \textbf{Indeed, our results indicate that Spontaneous real faces have greater intensity at the Upper of the face than in the Lower part}, which is confirmed by Park et al.~\cite{park:2020} work. 
On the other hand, Park found that the posed smiles were more intense on the Lower part of the face. However, this finding was not confirmed in our study. Indeed, in the present work, we observed that \textbf{CG faces presented such characteristics, i.e., the lower part of the face got higher intensity values than the upper part.}


\subsection{Smoothing Intensities in CG faces}
\label{smoothing_intensitiies_in_cg}

Analyzing the CG faces, transformed from Deep3D~\cite{deng2019accurate}, in both AU6 and AU12, comparing Figure~\ref{fig:comparative1} (a) with (b), and (c) with (d), it is noticeable the smoothing of the intensity medians, as well as the standard deviation and the variance. 
The intensity values for AU6 and AU12 in Posed faces decreased by approximately 70\% and 32\%, respectively, for both genders, while for Spontaneous faces, the smoothing rates for AU6 and AU12 were observed to reach as high as 80\% and 45\%, respectively, for both genders.

This observation was previously discussed by M{\"a}k{\"a}r{\"a}inen et al.~\cite{makarainen2014exaggerating}, who hypothesized that when a facial expression is directly copied from one domain to another, it may appear less intense on a more abstract face. Our results are in accordance with the authors once the AU intensity levels have been decreased when transferred to CG faces. 

The factor that may contribute to the smoothing observed in the data is the result (rendering) of the 3D reconstruction models, which still cannot achieve the desired realism. Here, we hypothesized that the used software may smooth the expression intensities. To verify that, we proceed with a case study evaluating two media characters using our methodology. Our goal is to evaluate the smoothness that happens when real faces are transferred to CG.

\subsection{Case Studies: Genius and She-Hulk}
\label{case_studies}

\begin{figure*}[t!]
    \centering
    \begin{subfigure}[b]{0.4\linewidth}        
        \centering
        \includegraphics[width=\linewidth]{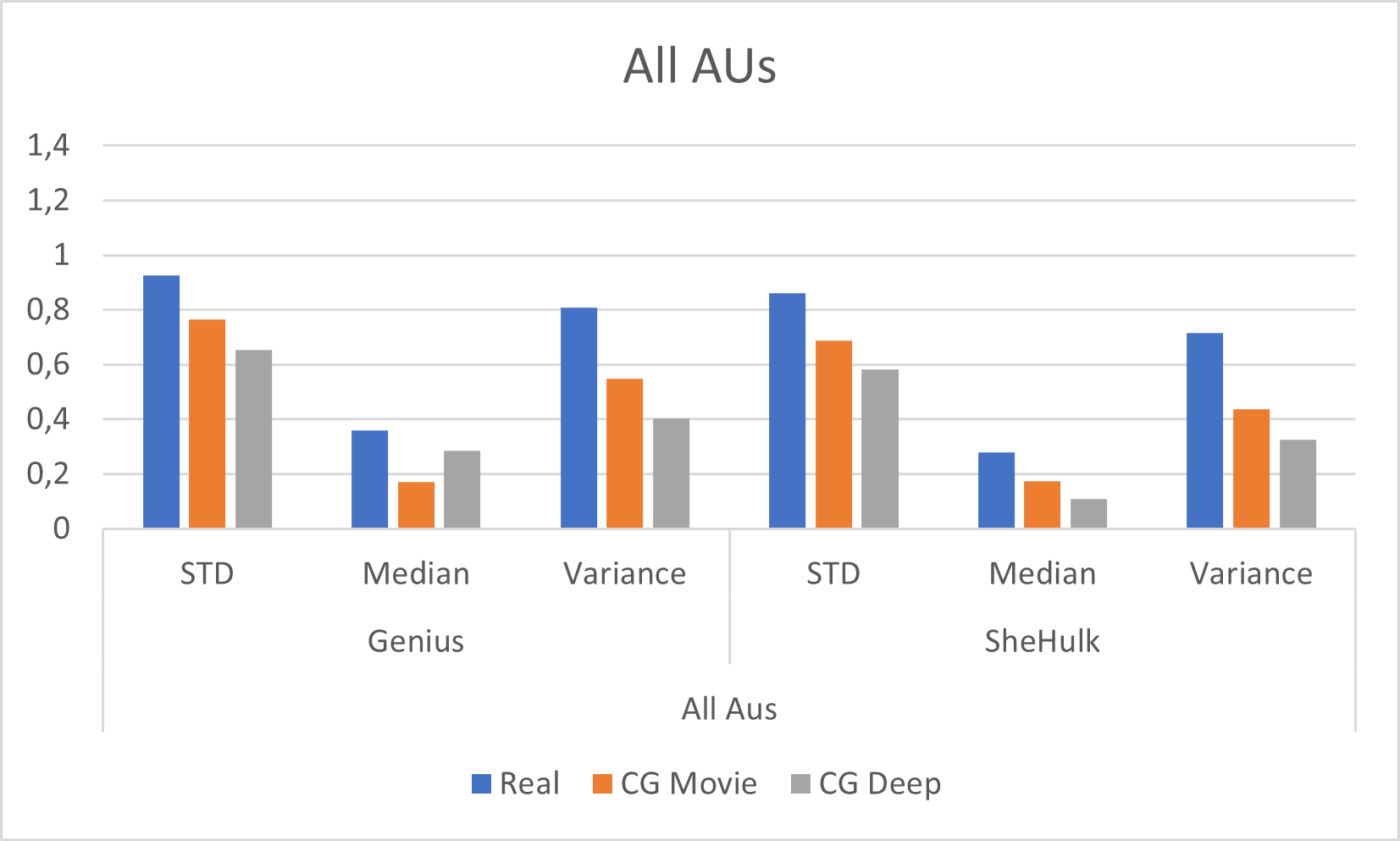}
        \caption{All AU Genius x She-Hulk}
        \label{fig:shehulk_genius_all_AU}
    \end{subfigure}
    \begin{subfigure}[b]{0.4\linewidth}        
        \centering
        \includegraphics[width=\linewidth]{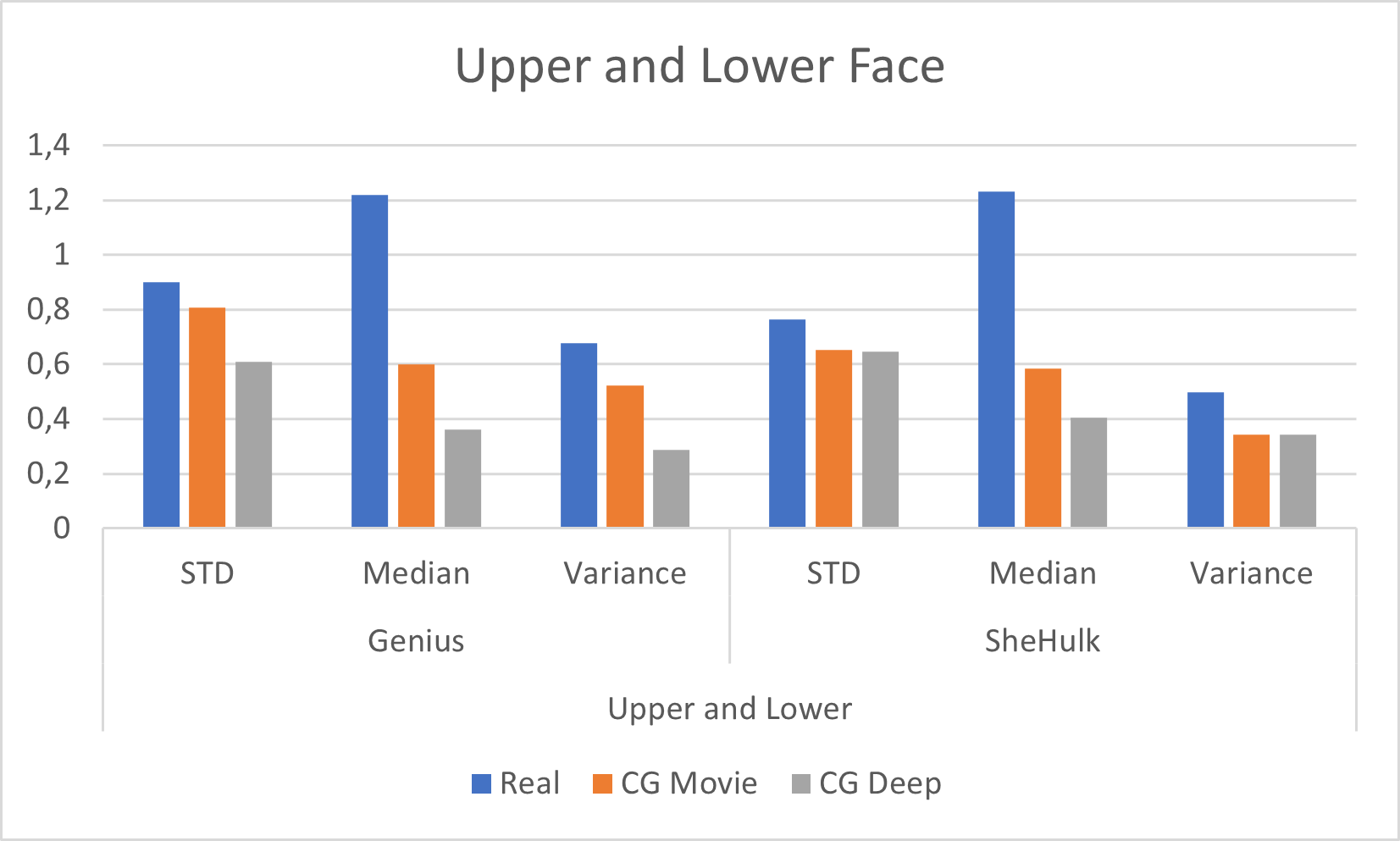}
        \caption{Upper and Lower Face (AU6+AU12) Genius x She-Hulk}
        \label{fig:shehulk_genius_top_and_bottom}
    \end{subfigure}

    \begin{subfigure}[b]{0.4\linewidth}        
        \centering
        \includegraphics[width=\linewidth]{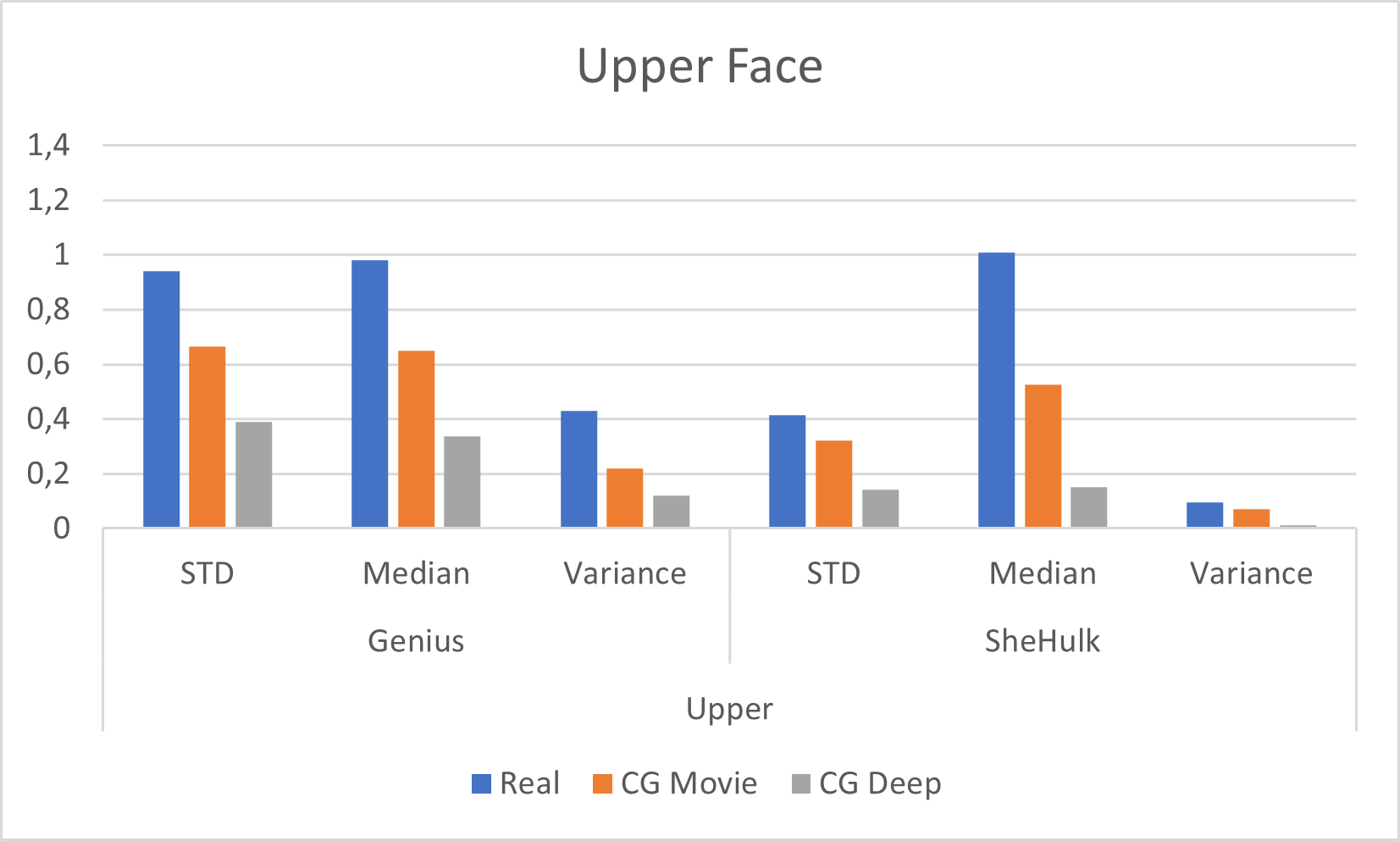}
        \caption{Upper Face (AU6) Genius x She-Hulk}
        \label{fig:shehulk_genius_top}
    \end{subfigure}
    \begin{subfigure}[b]{0.4\linewidth}        
        \centering
        \includegraphics[width=\linewidth]{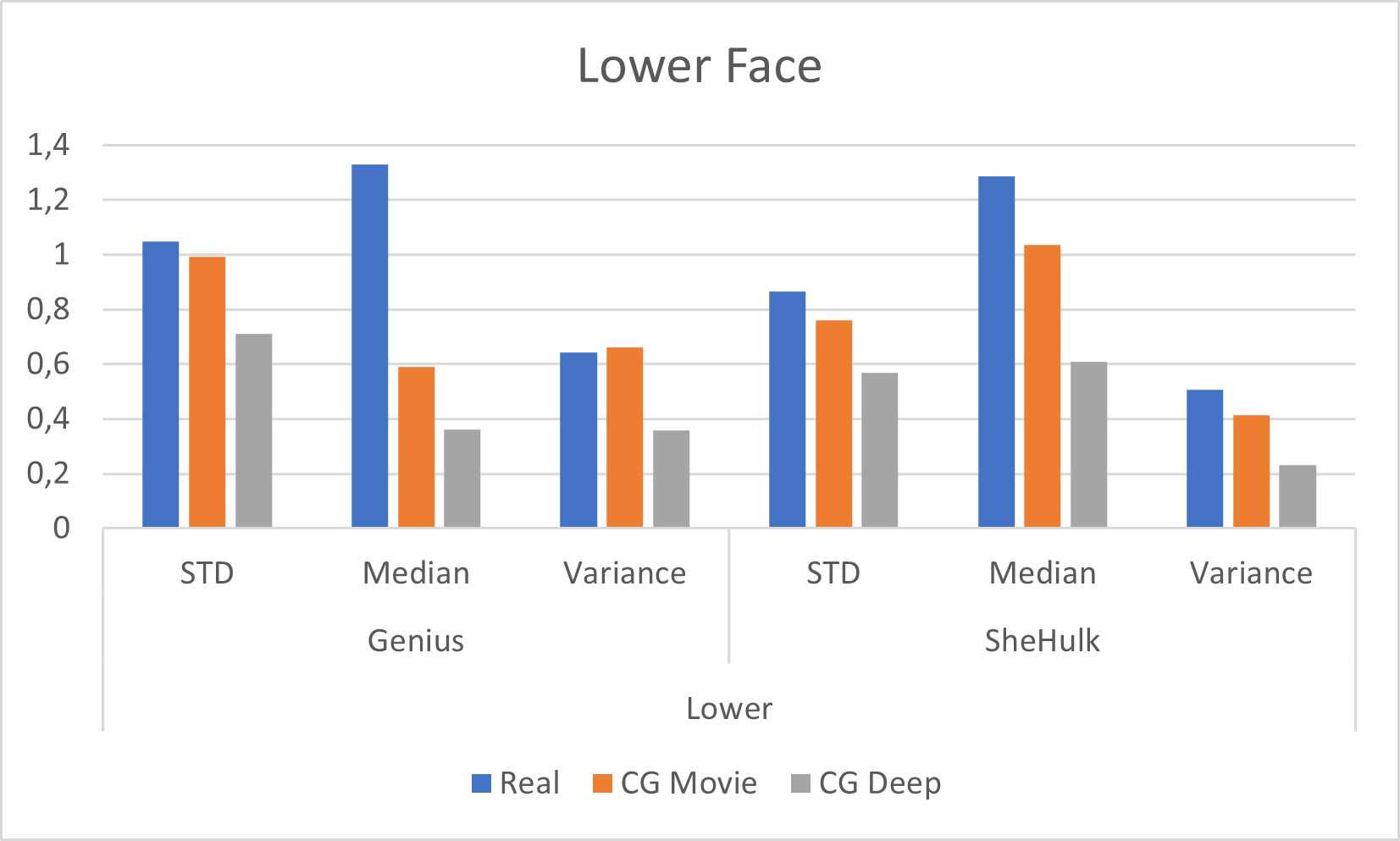}
        \caption{Lower Face (AU12) Genius x She-Hulk}
        \label{fig:shehulk_genius_bottom}
    \end{subfigure}
    
    \caption{The graphs show the standard deviation (STD), median and variance which are statistical measures used to evaluate the intensity values of AU referring to She-Hulk and Genius characters in Real face domains, CG Movie and CG generated in Deep3D~\cite{deng2019accurate}.}
    \label{fig:comparative2}
\end{figure*}

In this section, we evaluate what happened to the facial expressions when we transformed real faces to CG with the Deep3D tool~\cite{deng2019accurate} and compared it with the CG presented in the studied films. 

As shown in the last analysis, there is also a reduction of AU intensity values when compared Real x CG, which is easily observed in Fig~\ref{fig:shehulk_genius_all_AU}, Fig~\ref{fig:shehulk_genius_top_and_bottom}, Fig~\ref{fig:shehulk_genius_top} and Fig~\ref{fig:shehulk_genius_bottom}. M{\"a}k{\"a}r{\"a}inen et. al.~\cite{makarainen2014exaggerating}'s studies indicate that smoothing out facial expressions not only generalizes the emotion (being less specific) but also allows the post-processing artist greater freedom to enhance the CG's facial expression of emotion. In this case, the representation of the virtual face is less faithful to real movement but allows extra artistic work. In our tests, it is easy to see that intensity values reduce in the median, standard deviation, and variance.

The standard deviation (STD) and the variance presented in Figure~\ref{fig:shehulk_genius_all_AU}, Fig~\ref{fig:shehulk_genius_top_and_bottom}, Fig~\ref{fig:shehulk_genius_top} and Fig~\ref{fig:shehulk_genius_bottom} shows that for the Genius and She-Hulk faces values are always higher in Real faces, then in CG Movies and lastly in the faces generated using Deep3D tool~\cite{deng2019accurate}. So, in addition to the high values of the median, there is more variation also in the Real faces of the actors, confirming that independent of the software we used to transform the domain of images from Real faces to CG, there is a smoothness in the AU intensities.

Still, in the study by M{\"a}k{\"a}r{\"a}inen et al.~\cite{makarainen2014exaggerating}, the exaggeration approach is recognized as one of the most well-known animation principles. According to studies by Johnston et al. ~\cite{thomas1995illusion}, its use is relevant to the perception of emotional expressions. These fundamental principles of animation have been applied in 3D computer animation, as indicated by Lasseter et al. ~\cite{lasseter1987principles} and in other applications such as robotics.

Our studies indicate that while faces in the CG domain that corresponds to Real faces tend to be smoothed, media characters in movies, that artists have postprocessing worked on their animation, can amplify the final AU intensities.

\begin{table} 
    \small
    \centering 

        \begin{tabular}{l c c  r} 
        \hline
            \textbf{Factor} & \textbf{Sum of squares} & \textbf{F} &\textbf{$f$-value} 
            \\ [0.5ex] 
         \hline 
            
            \multicolumn{4}{l}{(a) two-way ANOVA on AU6 intensity} \\ \hline
            
            Gender & 41.70979 & 215.86032 & \textbf{$<$ 1.3627e-44} 
            \\
            Domain & 93.55025 & 242.07490 & \textbf{$<$ 3.1802e-87} 
            \\
            Gender x Domain & 0.50547 & 1.30797 & 0.2708 
            \\ \hline

            \multicolumn{4}{l}{(b) two-way ANOVA on AU12 intensity} \\ \hline
         
            Gender & 27.58650 & 68.00717 & \textbf{$<$ 4.8163e-16} 
            \\
            Domain & 39.48873 & 48.67447 & \textbf{$<$ 6.0499e-21} 
            \\ 
            Gender x Domain & 18.20199 & 22.43608 & \textbf{$<$ 2.8709e-10} 
            \\ \hline
            
        \end{tabular}
    
    \caption{Two-way (Gender x Domain) ANOVA of the main effect and the interaction effect on (a) AU6 intensity and (b) AU12 intensity. The significant variables and their corresponding $p$-values are highlighted. $p$-value < 0.05 is considered as statistically significant.} 
    \label{tab:anova_2} 
\end{table}

Table~\ref{tab:anova_2} provides an overview of the two variables and their effects on the intensity of AU6 and AU12. Independent variables include Gender (Men and Women) and Domain (Real, CG Movie, CG Deep3D). A two-way analysis of variance (ANOVA) (Gender × Domain) is performed on the intensities AU6 and AU12 separately (Table~\ref{tab:anova_2}). 
The two-way ANOVA (Table~\ref{tab:anova_2}) shows that Gender (F=215.86032, $f$-value=1.3627e-44) and Domain (F=242.07490, $f$-value=3.1802e-87) have a significant effect on the intensity of AU6. This no longer occurs for the Gender x Domain interaction (F=1.30797, $f$-value=0.2708), as it is not significant.  

    
    
    

    


\begin{figure}[h] \centering
    \subfloat[Upper Face (AU6) Genius x She-Hulk]
    {\includegraphics[width=0.45\textwidth, keepaspectratio]{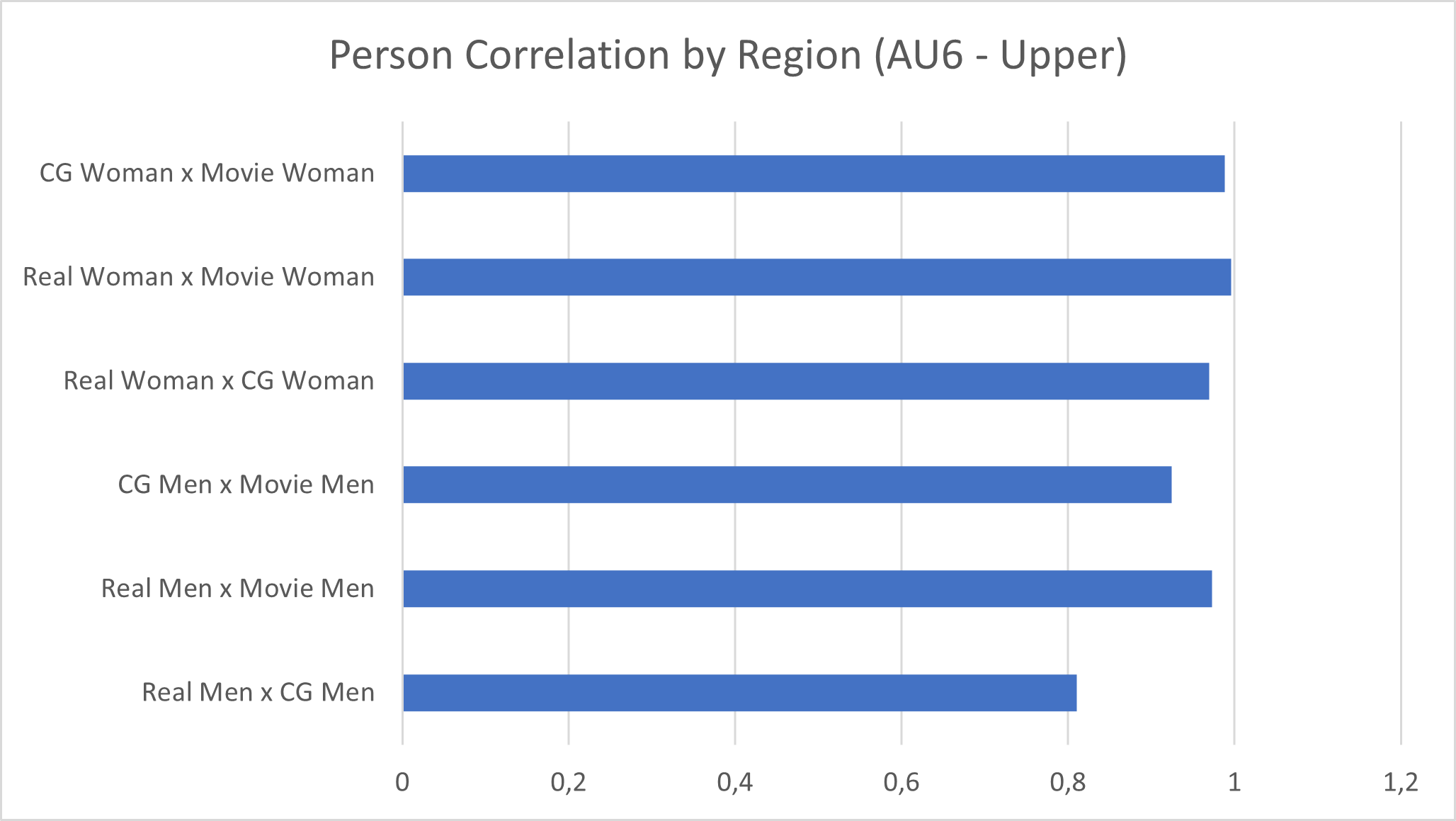}}
    \hspace{2mm}
    \subfloat[Lower Face (AU12) Genius x She-Hulk]
    {\includegraphics[width=0.45\textwidth, keepaspectratio]{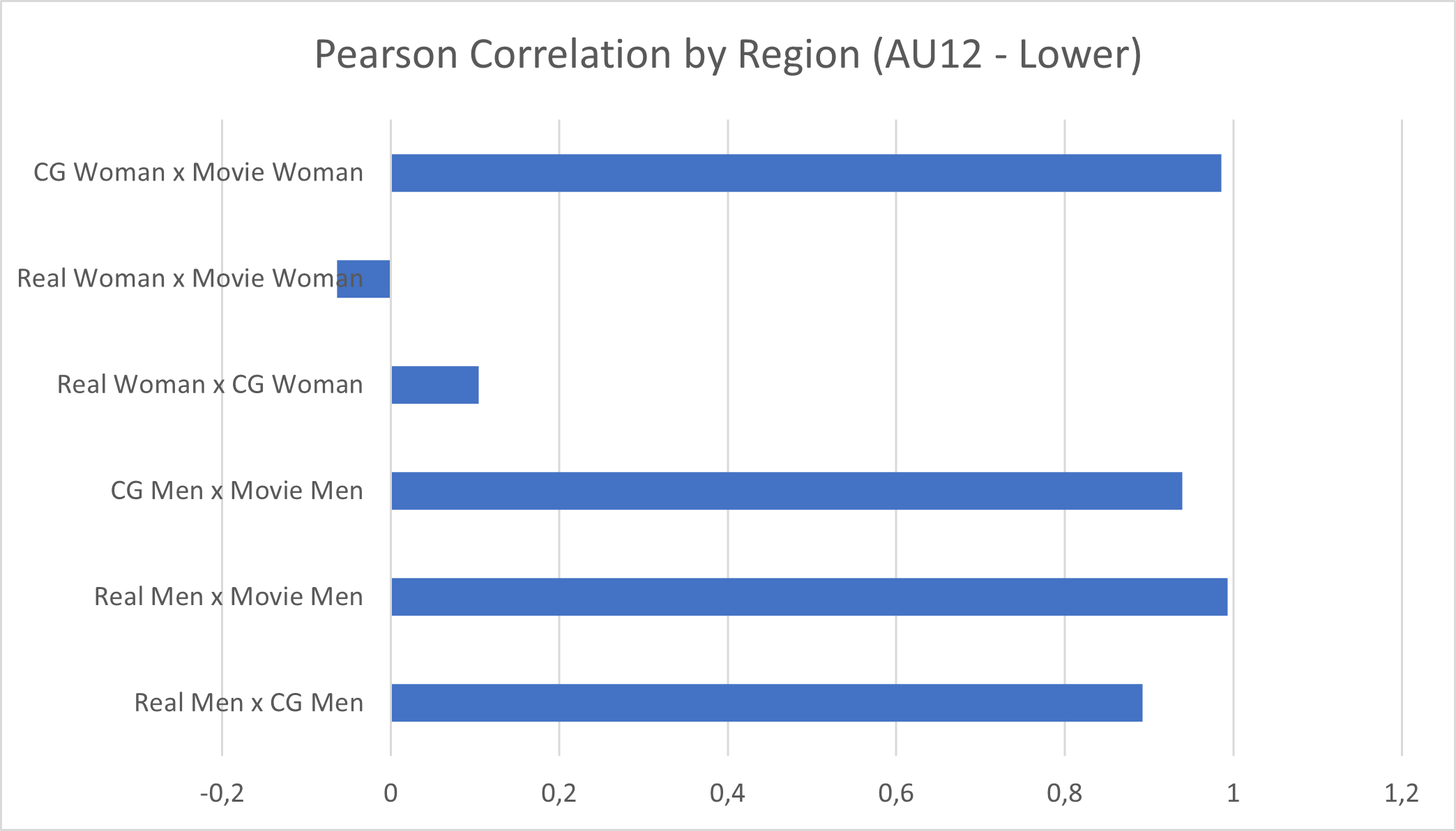}}
    \hspace{2mm}
    \subfloat[Upper and Lower Face (AU6 + AU12) Genius x She-Hulk]
    {\includegraphics[width=0.45\textwidth, keepaspectratio]{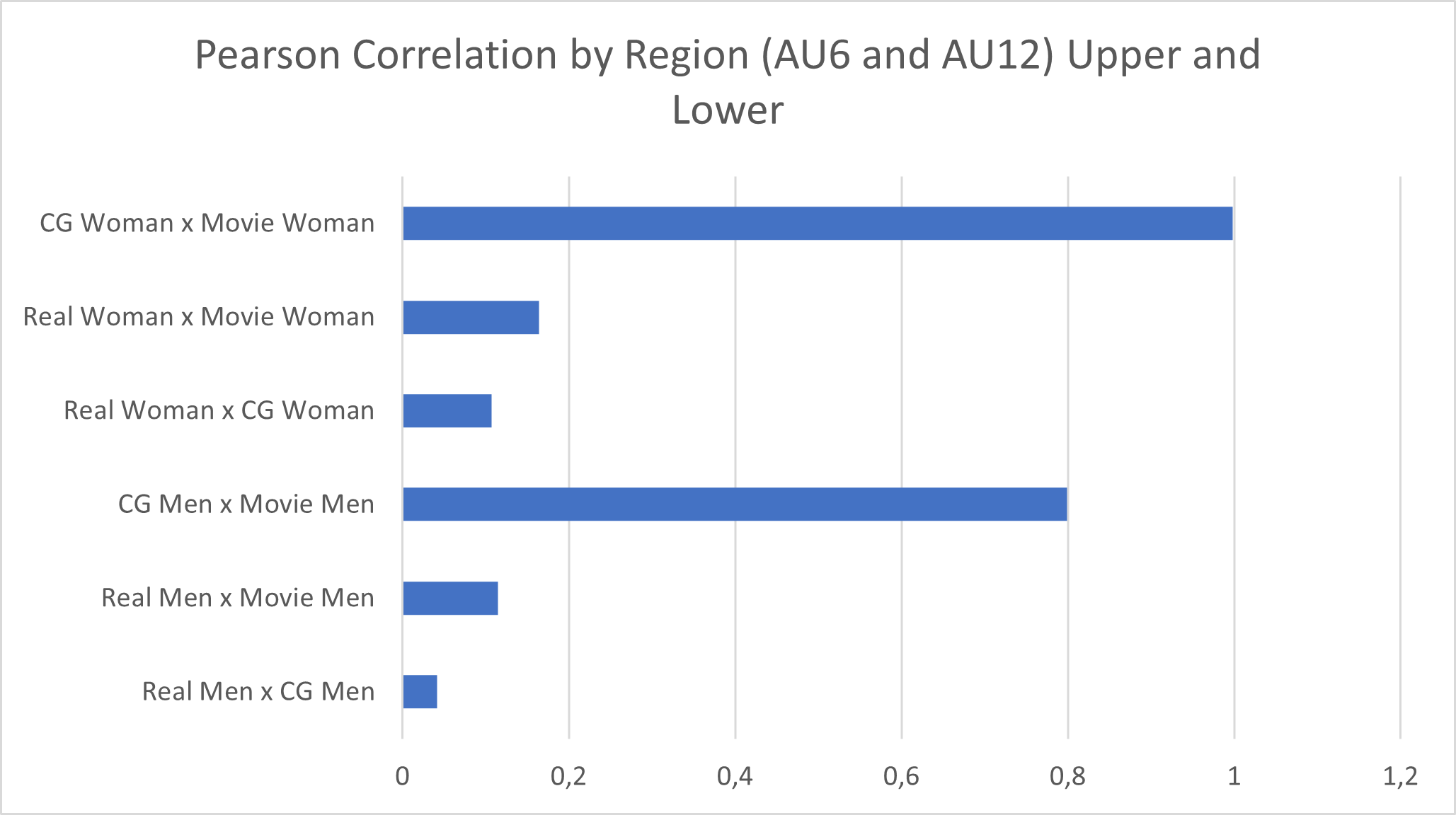}}
    \caption{Pearson’s correlations regarding the AU intensity values for AU6, AU12 and the both of them.}
    \label{fig:correlation_movies}
\end{figure}

Regarding the AU12 intensity values, the two-way ANOVA (Table~\ref{tab:anova_2}) shows that Gender (F=68.00717, $f$-value=4.8163e-16), Domain (F =48.67447, $f $-value=6.0499e-21) and the Gender x Domain interaction (F=22.43608, $f$-value=2.8709e-10) have a significant effect. For the domain, we can confirm this in Figure~\ref{fig:comparative2}(b). In the case of AU6, the ANOVA shows that the Gender x Domain interaction is not significant (F=1.30797, $f$-value=0.2708) and we can also confirm this through Figure~\ref{fig:correlation_movies}(a) because the correlations are all high. It is possible to verify in Figure~\ref{fig:correlation_movies}(b) that there is only a small correlation when the domain is Real and CG Movie/CG Deep3D for the Women, otherwise the correlation is very high.

Figure~\ref{fig:correlation_movies}(c), considering the intensities of the Upper and Lower AU, shows that there is a strong correlation between CG Deep3D tool~\cite{deng2019accurate} and CG Movie for She-Hulk and Genius. The opposite happens when the Pearson comparison is made between Real faces and corresponding CG Deep3D/CG movie faces. Considering only the intensity of the Upper part (AU6)~\ref{fig:correlation_movies}(a), we can see that all comparisons show high correlations, while the intensity of the Lower face (AU12)~\ref{fig:correlation_movies}(b) shows a low correlation in the comparisons between Real x CG Movie and for Real x CG Deep3D in the actress. 

\section{Final Considerations}
\label{FinalConsiderations}

This paper presented a study about Real and CG faces expressing Happiness emotion in Women/Men and Posed/Spontaneous faces. Additionally, we analyzed two media characters that are animated based on real actors' performances. This work first allowed us to generate datasets of real people having the following features: \textit{i)} Posed/Spontaneous, \textit{ii)} in CG or Real domains, generated with the Deep3D tool. In addition, 3 more datasets of faces of two known actors were generated, the first dataset being composed of real faces, the second using faces generated by Deep3D and the last one using animated CG images of the actors in their respective films, She-Hulk and Genius.

We performed a three-way and two-way ANOVA analysis to assess the significance of the Gender, Domain, and Dataset factors. The Pearson correlation between the factors was also performed and graphs were created to analyze the standard deviation, median, and variance of the AU' intensities values.

Our main findings are:
\begin{itemize}
    \item Upper part of the face expresses higher values of intensities than the Lower part, in Real faces (also according to Park et al~\cite{park:2020});
    \item AU intensities of Posed faces are always higher than Spontaneous ones;
    \item The intensities of AUs are higher in the Lower part of the face (detailed in Section~\ref{sec:Comparison}) in the CG faces than in the Upper part, as indicated in the study by Park et al~\cite{park:2020} for the Posed dataset;
    \item The Gender factor did not reveal significant differences (Table~\ref{tab:anova_1}) in the intensity of AU6 between Men and Women (Figure~\ref{fig:comparative1}) in both the Posed and Spontaneous datasets and in real/CG faces;
    \item Through the median, we can see the smoothing of the intensity of the AU6 and AU12 as discussed in Section~\ref{case_studies}, with respect to the media characters She-Hulk and Genius. The smoothing is higher in the domain of CG Deep3D and then CG Movie. This result is in accordance with the studies by M{\"a}k{\"a}r{\"a}inen et al. ~\cite{makarainen2014exaggerating}, which discusses the smoothing of facial expressions as a way to allow the artist to have greater freedom in post-processing the animations to enhance the facial expression of emotions of the CG characters; and finally
    \item We observed that the Upper part of the face better correlates Real and CG actors than the Lower part for posed datasets. On the other hand, the lower part of the face presents higher values of Pearson correlation when comparing Real and CG faces, in general.
\end{itemize}

In the present work, we aimed to answer the questions related to the main research question: Are we really transferring the actor's genuine emotions to the avatars?

\begin{itemize}
      \item \textbf{RQ1:} Are the real faces and their respective correspondences in CG similar considering the activated AUs (Lower or Upper parts of the face), the actors' gender (women or men), and the type of dataset (Spontaneous or Posed), considering only the happiness emotion?
      \subitem ANSWER: Real and CG faces are differently impacted by the Upper and Lower parts of the face. The actors' gender is insignificant in the difference, and AU intensities are always higher in Posed datasets (in real and CG faces).
       \item \textbf{RQ2:} Is there a smooth effect when transforming from Real to CG faces?
       \subitem ANSWER: Yes, there is a smoothing effect when we transfer from Real to CG faces. Furthermore, some patterns were observed in the transformed datasets, i.e., facial expressions of real humans are always more intense than CG faces.  
       \item \textbf{RQ3:} How the metrics defined in RQ1 work for existing animated characters from the movie industry?
       \subitem ANSWER: 
       Indeed, the same patterns observed in CG faces transformed using Deep3D were observed with CG actors manipulated by artists, e.g., in the cases of She-Hulk and Genius.
\end{itemize}

We verified that the intensity of AU6 and AU12 in the dataset (spontaneous and posed) suggests that the difference in intensity of these AUs concerning the actors' gender is very subtle, as we can see in Figure~\ref{fig:comparative1} (c) and (d ), contradicting the work by Fan et al.~\cite{fan2021demographic} who indicated that the female group had a higher AU12 intensity than the male group.
However, when observing the use case of She-Hulk and Genius in Figure~\ref{fig:comparative2}(d), we can see that there is a greater intensity of AU12 in She-Hulk compared to Genius, as proposed by Fan et al.~\cite{fan2021demographic}. We can also observe that the She-Hulk in CG Movie indicates an AU12 intensity much closer to the Real domain than the CG Deep3D. 

Finally, as per our assessments, we believe that we are not transferring the actor's genuine emotions to the avatars, using the reconstruction software. The transformation between Real and CG domains still needs tweaking, so the presence of the artist is still critical to exaggerating the facial expression in CG. However, we believe that even the artist still has limitations to transfer the actor's genuine emotion to the avatars.

In the future, we intend to explore other emotions and mixtures of emotions, adding them to our real and CG datasets. In this way, we can suggest an analysis method for the transformation from real to CG that can make it more reliable with human facial expressions. Indeed, we intend to provide a methodology that allows us to analyze real and virtual faces, providing information about which part of the face should be exaggerated to better reflect real faces' motion. Another aspect we want to investigate is the recognition of action units and their intensity values using models trained with annotated CG faces.

\section{Acknowledgments}

We would like to thank CNPq and CAPES for partially supporting this work.

\bibliographystyle{unsrt}
\bibliography{acmart}

\begin{thebibliography}{10}

\bibitem{ekman1969repertoire}
Paul Ekman and Wallace~V Friesen.
\newblock The repertoire of nonverbal behavior: Categories, origins, usage, and coding.
\newblock {\em semiotica}, 1(1):49--98, 1969.

\bibitem{Lance1990}
Lance Williams.
\newblock Performance-driven facial animation.
\newblock {\em SIGGRAPH Comput. Graph.}, 24(4):235–242, sep 1990.

\bibitem{Barros2019}
Jilliam~María Díaz~Barros, Vladislav Golyanik, Kiran Varanasi, and Didier Stricker.
\newblock Face it!: A pipeline for real-time performance-driven facial animation.
\newblock In {\em 2019 IEEE International Conference on Image Processing (ICIP)}, pages 2209--2213, 2019.

\bibitem{barrett2019emotional}
Lisa~Feldman Barrett, Ralph Adolphs, Stacy Marsella, Aleix~M Martinez, and Seth~D Pollak.
\newblock Emotional expressions reconsidered: Challenges to inferring emotion from human facial movements.
\newblock {\em Psychological science in the public interest}, 20(1):1--68, 2019.

\bibitem{park:2020}
Seho Park, Kunyoung Lee, Jae-A Lim, Hyunwoong Ko, Taehoon Kim, Jung-In Lee, Hakrim Kim, Seong-Jae Han, Jeong-Shim Kim, Soowon Park, Jun-Young Lee, and Eui~Chul Lee.
\newblock Differences in facial expressions between spontaneous and posed smiles: Automated method by action units and three-dimensional facial landmarks.
\newblock {\em Sensors}, 20:1199, 02 2020.

\bibitem{ekman1978}
P.~Ekman and W.V. Friesen.
\newblock {\em Facial Action Coding System: Manual}.
\newblock Number v. 1-2. Consulting Psychologists Press, 1978.

\bibitem{makarainen2014exaggerating}
Meeri M{\"a}k{\"a}r{\"a}inen, Jari K{\"a}tsyri, and Tapio Takala.
\newblock Exaggerating facial expressions: A way to intensify emotion or a way to the uncanny valley?
\newblock {\em Cognitive Computation}, 6:708--721, 2014.

\bibitem{deng2019accurate}
Yu~Deng, Jiaolong Yang, Sicheng Xu, Dong Chen, Yunde Jia, and Xin Tong.
\newblock Accurate 3d face reconstruction with weakly-supervised learning: From single image to image set.
\newblock In {\em IEEE Computer Vision and Pattern Recognition Workshops}, 2019.

\bibitem{DECA:Siggraph2021}
Yao Feng, Haiwen Feng, Michael~J. Black, and Timo Bolkart.
\newblock Learning an animatable detailed {3D} face model from in-the-wild images.
\newblock volume~40, 2021.

\bibitem{EMOCA:CVPR:2021}
Radek Danecek, Michael~J. Black, and Timo Bolkart.
\newblock {EMOCA}: {E}motion driven monocular face capture and animation.
\newblock In {\em Conference on Computer Vision and Pattern Recognition (CVPR)}, pages 20311--20322, 2022.

\bibitem{amos2016openface}
Brandon Amos, Bartosz Ludwiczuk, and Mahadev Satyanarayanan.
\newblock Openface: A general-purpose face recognition library with mobile applications.
\newblock Technical report, CMU-CS-16-118, CMU School of Computer Science, 2016.

\bibitem{park2020differences}
Seho Park, Kunyoung Lee, Jae-A Lim, Hyunwoong Ko, Taehoon Kim, Jung-In Lee, Hakrim Kim, Seong-Jae Han, Jeong-Shim Kim, Soowon Park, et~al.
\newblock Differences in facial expressions between spontaneous and posed smiles: Automated method by action units and three-dimensional facial landmarks.
\newblock {\em Sensors}, 20(4):1199, 2020.

\bibitem{fan2021demographic}
Yingruo Fan, Jacqueline~CK Lam, and Victor~OK Li.
\newblock Demographic effects on facial emotion expression: an interdisciplinary investigation of the facial action units of happiness.
\newblock {\em Scientific reports}, 11(1):1--11, 2021.

\bibitem{sohre2017data}
Nick Sohre, Moses Adeagbo, and Stephen Guy.
\newblock Data-driven variation for virtual facial expressions.
\newblock 2017.

\bibitem{motley1988facial}
Michael~T Motley and Carl~T Camden.
\newblock Facial expression of emotion: A comparison of posed expressions versus spontaneous expressions in an interpersonal communication setting.
\newblock {\em Western Journal of Communication (includes Communication Reports)}, 52(1):1--22, 1988.

\bibitem{Ma:2015}
Debbie Ma, Joshua Correll, and Bernd Wittenbrink.
\newblock The chicago face database: A free stimulus set of faces and norming data.
\newblock {\em Behavior research methods}, 47, 01 2015.

\bibitem{Lakshmi2021TheIF}
Anjana Lakshmi, Bernd Wittenbrink, Joshua Correll, and Debbie~S. Ma.
\newblock The india face set: International and cultural boundaries impact face impressions and perceptions of category membership.
\newblock {\em Frontiers in Psychology}, 12, 2021.

\bibitem{THOMAZ2010902}
Carlos~Eduardo Thomaz and Gilson~Antonio Giraldi.
\newblock A new ranking method for principal components analysis and its application to face image analysis.
\newblock {\em Image and Vision Computing}, 28(6):902--913, 2010.

\bibitem{debruine2017face}
Lisa DeBruine and Benedict Jones.
\newblock Face research lab london set.
\newblock {\em Psychol. Methodol. Des. Anal}, 2017.

\bibitem{peres2021towards}
Vitor Miguel~Xavier Peres and Soraia~Raupp Musse.
\newblock Towards the creation of spontaneous datasets based on youtube reaction videos.
\newblock In George Bebis, Vassilis Athitsos, Tong Yan, Manfred Lau, Frederick Li, Conglei Shi, Xiaoru Yuan, Christos Mousas, and Gerd Bruder, editors, {\em Advances in Visual Computing}, pages 203--215, Cham, 2021. Springer International Publishing.

\bibitem{disfa:2013}
S.~M. {Mavadati}, M.~H. {Mahoor}, K.~{Bartlett}, P.~{Trinh}, and J.~F. {Cohn}.
\newblock Disfa: A spontaneous facial action intensity database.
\newblock {\em IEEE Transactions on Affective Computing}, 4(2):151--160, 2013.

\bibitem{pyfeat:2021}
Jin~Hyun Cheong, Tiankang Xie, Sophie Byrne, and Luke~J. Chang.
\newblock Py-feat: Python facial expression analysis toolbox.
\newblock {\em CoRR}, abs/2104.03509, 2021.

\bibitem{Keltner:2019}
Dacher Keltner, Disa Sauter, Jessica Tracy, and Alan Cowen.
\newblock Emotional expression: Advances in basic emotion theory.
\newblock {\em Journal of Nonverbal Behavior}, 43, 06 2019.

\bibitem{Ghayoumi}
Mehdi Ghayoumi and Arvind Bansal.
\newblock Unifying geometric features and facial action units for improved performance of facial expression analysis.
\newblock 06 2016.

\bibitem{8265507}
Amir Zadeh, Yao~Chong Lim, Tadas Baltrušaitis, and Louis-Philippe Morency.
\newblock Convolutional experts constrained local model for 3d facial landmark detection.
\newblock In {\em 2017 IEEE International Conference on Computer Vision Workshops (ICCVW)}, pages 2519--2528, 2017.

\bibitem{7553523}
Kaipeng Zhang, Zhanpeng Zhang, Zhifeng Li, and Yu~Qiao.
\newblock Joint face detection and alignment using multitask cascaded convolutional networks.
\newblock {\em IEEE Signal Processing Letters}, 23(10):1499--1503, 2016.

\bibitem{cai2021multi}
Xingjuan Cai, Yihao Cao, Yeqing Ren, Zhihua Cui, and Wensheng Zhang.
\newblock Multi-objective evolutionary 3d face reconstruction based on improved encoder--decoder network.
\newblock {\em Information Sciences}, 581:233--248, 2021.

\bibitem{lin2020towards}
Jiangke Lin, Yi~Yuan, Tianjia Shao, and Kun Zhou.
\newblock Towards high-fidelity 3d face reconstruction from in-the-wild images using graph convolutional networks.
\newblock In {\em Proceedings of the IEEE/CVF Conference on Computer Vision and Pattern Recognition}, pages 5891--5900, 2020.

\bibitem{nilsson2020understanding}
Jim Nilsson and Tomas Akenine-Möller.
\newblock Understanding ssim, 2020.

\bibitem{1284395}
Zhou Wang, A.C. Bovik, H.R. Sheikh, and E.P. Simoncelli.
\newblock Image quality assessment: from error visibility to structural similarity.
\newblock {\em IEEE Transactions on Image Processing}, 13(4):600--612, 2004.

\bibitem{thomas1995illusion}
Frank Thomas, Ollie Johnston, and Frank Thomas.
\newblock {\em The illusion of life: Disney animation}.
\newblock Hyperion New York, 1995.

\bibitem{lasseter1987principles}
John Lasseter.
\newblock Principles of traditional animation applied to 3d computer animation.
\newblock In {\em Proceedings of the 14th annual conference on Computer graphics and interactive techniques}, pages 35--44, 1987.

\end{thebibliography}

\end{document}